\documentclass[runningheads]{llncs}

\usepackage{eccv}

\usepackage{eccvabbrv}

\usepackage{graphicx}
\usepackage{booktabs}

\usepackage[accsupp]{axessibility}  

\usepackage{multirow, multicol} 
\usepackage{wrapfig}

\usepackage{dsfont}
\usepackage{amsmath,amssymb,amsfonts}

\usepackage{hyperref}

\usepackage{orcidlink}

\begin{document}
\title{PARE-Net: Position-Aware Rotation-Equivariant Networks for Robust Point Cloud Registration} 

\titlerunning{PARE-Net}

\author{Runzhao Yao\orcidlink{0000-0002-7106-9393} \and
Shaoyi Du\orcidlink{0000-0002-7092-0596} \thanks{Corresponding author: dushaoyi@xjtu.edu.cn}\and
Wenting Cui\orcidlink{0000-0003-2047-2091} \and
Canhui Tang\orcidlink{0009-0003-6797-3480} \and
Chengwu Yang\orcidlink{0009-0001-2438-5668}}

\authorrunning{R. Yao et al.}

\institute{National Key Laboratory of Human-Machine Hybrid Augmented Intelligence, National Engineering Research Center for Visual Information and Applications, and Institute of Artificial Intelligence and Robotics, Xi'an Jiaotong University, China}

\maketitle

\begin{abstract}
Learning rotation-invariant distinctive features is a fundamental requirement for point cloud registration. Existing methods often use rotation-sensitive networks to extract features, while employing rotation augmentation to learn an approximate invariant mapping rudely. This makes networks fragile to rotations, overweight, and hinders the distinctiveness of features. To tackle these problems, we propose a novel position-aware rotation-equivariant network, for efficient, light-weighted, and robust registration. The network can provide a strong model inductive bias to learn rotation-equivariant/invariant features, thus addressing the aforementioned limitations. To further improve the distinctiveness of descriptors, we propose a position-aware convolution, which can better learn spatial information of local structures. Moreover, we also propose a feature-based hypothesis proposer. It leverages rotation-equivariant features that encode fine-grained structure orientations to generate reliable model hypotheses. Each correspondence can generate a hypothesis, thus it is more efficient than classic estimators that require multiple reliable correspondences. Accordingly, a contrastive rotation loss is presented to enhance the robustness of rotation-equivariant features against data degradation. Extensive experiments on indoor and outdoor datasets demonstrate that our method significantly outperforms the SOTA methods in terms of registration recall while being lightweight and keeping a fast speed. Moreover, experiments on rotated datasets demonstrate its robustness against rotation variations. Code is available at \hyperlink{https://github.cYILFLZ-ZIGAGRom/yaorz97/PARENet}{https://github.com/yaorz97/PARENet}. 
\end{abstract}

\section{Introduction}
\label{sec:intro}

Point cloud registration, estimating a rigid transformation to align two point clouds that are partially overlapped, is fundamental research in 3D computer vision. It is widely applied in autonomous driving \cite{autodrive1,autodrive2}, robot localization \cite{robotics}, and 3D reconstruction \cite{zhang2020deep}. Feature-based point cloud registration framework is widely studied, which mainly involves descriptor extraction \cite{d3feat, predator,CoFiNet,qin2022geometric} and robust transformation estimation \cite{bai2021pointdsc,chen2022sc2, hunter}. One inherent contradiction in point cloud registration is the pose variation of point clouds and the desired invariance of their descriptors. To extract such rotation-invariant descriptors, plenty of approaches have been proposed that can be roughly categorized into patch-wise and scene-wise feature extractors.
\begin{figure}
    \centering
    \includegraphics[width=0.495\textwidth]{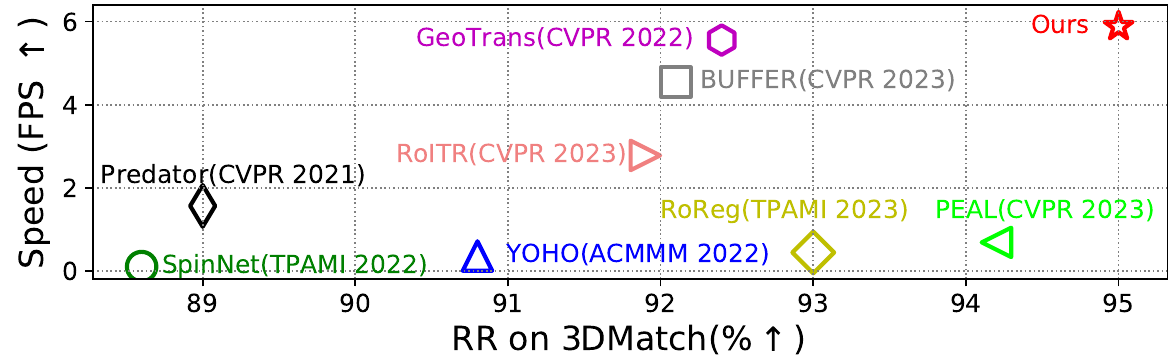}
    \includegraphics[width=0.495\textwidth]{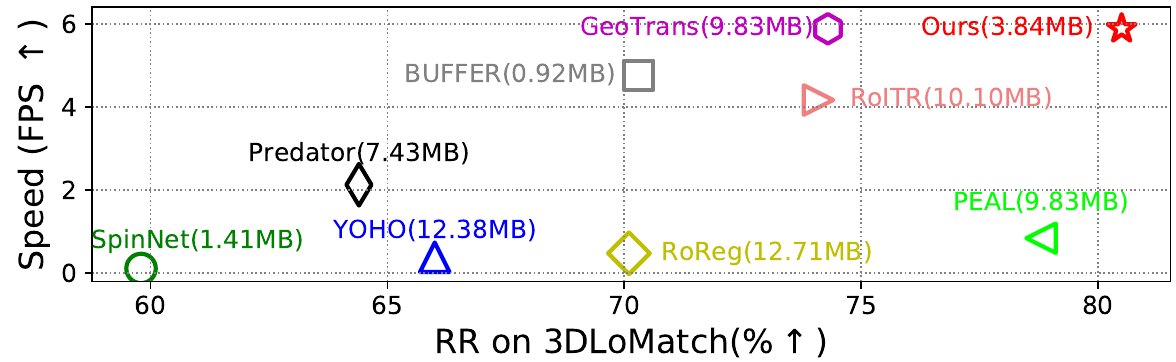}
    \caption{Experimental results on 3DMatch and 3DLoMatch. Our method significantly outperforms state-of-the-art methods w.r.t. registration recall 
(RR) while maintaining fast speed and lightweight. }
    \label{fig:teaser}
\end{figure}

 Patch-wise feature extractors \cite{ppfnet1,ppfnet2,spinnet} learn geometric features from local patches of point clouds and typically employ data preprocessing techniques to achieve rotation-invariance. Two common approaches include: 1) extracting rotation invariant statistics as initial representation from local patches, such as distance, angle, and point pair features (PPF \cite{ppfnet1,ppfnet2}); and 2) estimating a local reference frame (LRF\cite{spinnet}) to normalize poses of local patches. The former approach suffers from information loss, while the latter approach is susceptible to data occlusion and noise, lacking robustness. Moreover, these extractors are generally computationally expensive because all patches of a point cloud are processed individually, without sharing intermediate representations \cite{fcgf}.

 Scene-wise feature extractors \cite{fcgf, d3feat,predator,CoFiNet,qin2022geometric} learn dense feature descriptors for every point in the point cloud, and most of them employ extensive rotation augmentation in the training process to rudely enforce inherently rotation-sensitive networks (e.g., KPConv \cite{kpconv}) to learn rotation-invariant mappings. These methods have three shortcomings. Firstly, models are fragile to unseen rotations because it is difficult to cover the SO(3) space in the training process \cite{roitr}. Secondly, learning rotation-invariant mappings requires more network parameters. Thirdly, these networks are limited to generating discriminative feature descriptors as they cannot focus solely on learning structural distinctiveness. 

In this paper, we propose a novel registration network, PARE-Net, whose core is to fully exploit the advantages of rotation-equivariant networks in both feature extraction and transformation estimation, thereby achieving a lightweight, efficient, and robust registration method, as shown in Fig. \ref{fig:teaser}.

First of all, we propose a \textbf{P}osition-\textbf{A}ware \textbf{R}otation-\textbf{E}quivariant \textbf{Conv}olution (PARE-Conv) that enhances the representational power of networks for better distinctive feature learning. Vector Neuron (VN) \cite{vn} builds a general rotation-equivariant network framework. However, its representative power is limited as it ignores the importance of spatial localization property of local points and utilizes multi-layer perceptrons (MLPs) to crudely learn spatial encodings. Instead, PARE-Conv is proposed to explicitly leverage the spatial information of point clouds while guaranteeing the rotation equivariant property. Specifically, we first define a set of shadow kernel points equipped with learnable weights. Then, we introduce a correlation network that predicts rotation-invariant correlations between points and shadow kernels according to rotation-equivariant spatial features of points. Therefore, the geometric structures of point clouds are well captured. Finally, the correlation scores are subsequently leveraged to dynamically assemble kernel weights for rotation-equivariant convolution. As a result, PARE-Conv inherently provides a strong model bias to learn rotation-equivariant/invariant features, thus addressing the aforementioned problems raised by rotation-sensitive networks. Moreover, it can learn the spatial structure of point clouds better, facilitating to extraction of more distinctive descriptors.

\begin{wrapfigure}{r}{0.30\textwidth}
    \centering
    \includegraphics[width=.30\textwidth]{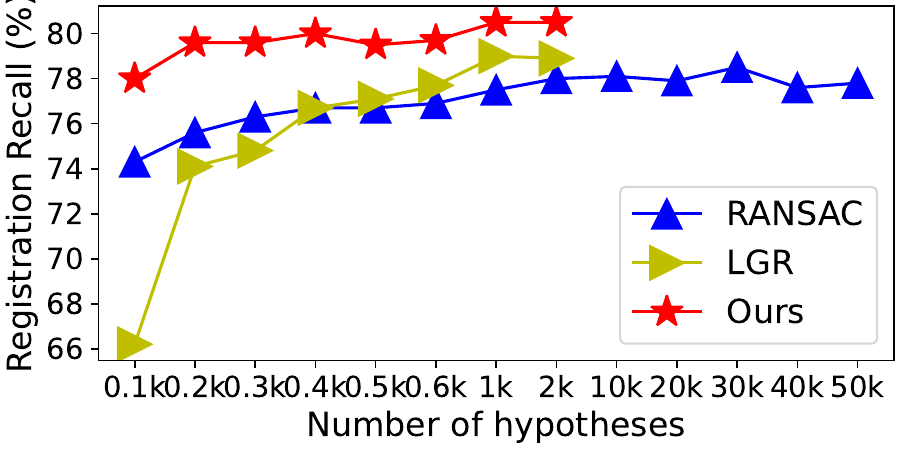}
    \caption{Comparison of our feature-based hypothesis proposer, RANSAC and LGR on 3DLoMatch. }
    \label{fig:hypo}
 \end{wrapfigure}

Secondly, we propose a feature-based hypothesis proposer that can efficiently generate reliable solution hypotheses. Once obtaining correspondences by feature matching, rotation-equivariant features encoding fine-grained structure information are used to explicitly generate multiple hypotheses. This way can alleviate the orientation ambiguity stemming from optimizing distance errors of correspondences. Moreover, to make the equivariant features more robust against data degradation, we present a contrastive rotation loss inspired by metric learning \cite{kaya2019deep, fcgf}, facilitating more robust pose estimation. As shown in Fig. \ref{fig:hypo}, our method can achieve significantly higher registration recall compared to RANSAC \cite{ransac} and LGR (presented in GeoTrans \cite{qin2022geometric}) when varying the number of hypotheses, demonstrating our method is more robust than classical estimators. Because they require multiple correspondences and the probability of sampling all correct correspondences decreases rapidly as the number of used correspondences increases.
Overall, our main contributions are:
\begin{itemize}
    \item A lightweight, efficient, and robust point cloud registration method that is achieved by fully exploiting the advantages of
 rotation-equivariant networks in feature extraction and pose estimation.
    \item A position-aware rotation-equivariant convolution that can better leverage spatial information, enabling the extraction of discriminative descriptors.
    \item A feature-based hypothesis proposer that can efficiently generate multiple reliable hypotheses and a contrastive rotation loss makes the rotation-equivariant features more robust to data degradation.
\end{itemize}

\section{Related Work}
\label{sec:related_work}
\textbf{Patch-wise feature extractors}. In the early stage, most learning-based feature descriptors \cite{zeng20173dmatch, khoury2017learning, ppfnet1, ppfnet2, gojcic2019perfect, spinnet} were patch-wise extractors, primarily due to limitations in point cloud convolution techniques, which could not handle large-scale point clouds effectively. During this period, the works drew inspiration from traditional feature descriptors \cite{guo2016comprehensive} and focused on designing features with rotation-invariance. PPFNet \cite{ppfnet1} and PPFFoldNet \cite{ppfnet2} extracted rotation-invariant PPFs as initial features and further learned high-dimensional features. PerfectMatch \cite{gojcic2019perfect} and SpinNet \cite{spinnet} utilized LRF to normalize the pose of local patches and then employed networks to learn the density/structural features of patches. These methods are primarily limited by their computational efficiency.

\textbf{Scene-wise feature extractors} \cite{fcgf, d3feat, predator, CoFiNet, lepard, regtr, qin2022geometric, peal, roitr} have become the mainstream research in recent years, due to their high efficiency. These methods mainly achieved rotation invariance through data augmentation and aim to tackle the low-overlapped problem. Predator \cite{predator} designed an overlap-attention module to detect overlap regions between two point clouds. CoFiNet \cite{CoFiNet} and GeoTrans \cite{qin2022geometric} presented a coarse-to-fine matching scheme, and GeoTrans introduced a geometric transformer module that significantly increased the matching accuracy by injecting rotation-invariant position information. PEAL \cite{peal} presented an explicit one-way attention module that iteratively leveraged overlap prior to relieving feature ambiguity while bringing expensive computation costs.
Moreover, some methods with rotation invariant/equivariant designs were proposed. YOHO \cite{yoho} and RoReg \cite{roreg} exploited the group convolution technique to extract rotation-invariant/equivariant group features, which are extremely computation expensive. Because they need to extract point features under 60 different poses and perform convolutions on massive point features. RoITR \cite{roitr} used PPF to design an invariant convolution and presented a rotation invariant transformer to extract distinctive features. BUFFER \cite{buffer} combined scene-wise and patch-wise feature extractors to achieve faster alignment with great generalization ability. It first leveraged a rotation-equivariant network built on vanilla VN to extract invariant and equivariant features for keypoints and orientations prediction. Then, it also used SpinNet to extract patch-wise features due to the limited representative ability of vanilla VN.

\textbf{Robust transformation estimators}. Since the estimated correspondences are usually contaminated by outliers, a robust transformation estimator is necessary. RANSAC \cite{ransac} is the most widely used method due to its simplicity, but it typically exhibits lower accuracy and requires plenty of iterations for convergence. Recently, some learning-based methods \cite{bai2021pointdsc, lee2021deep, chen2022sc2, hunter} are proposed, whose core idea is to design networks to learn spatial consistency between correspondences, enabling efficient inlier identification and reliable hypothesis generation. We compare our approach with these methods in Appendix F.1.

\section{Method}

\begin{figure}
    \centering
    \includegraphics[width=0.95\textwidth]{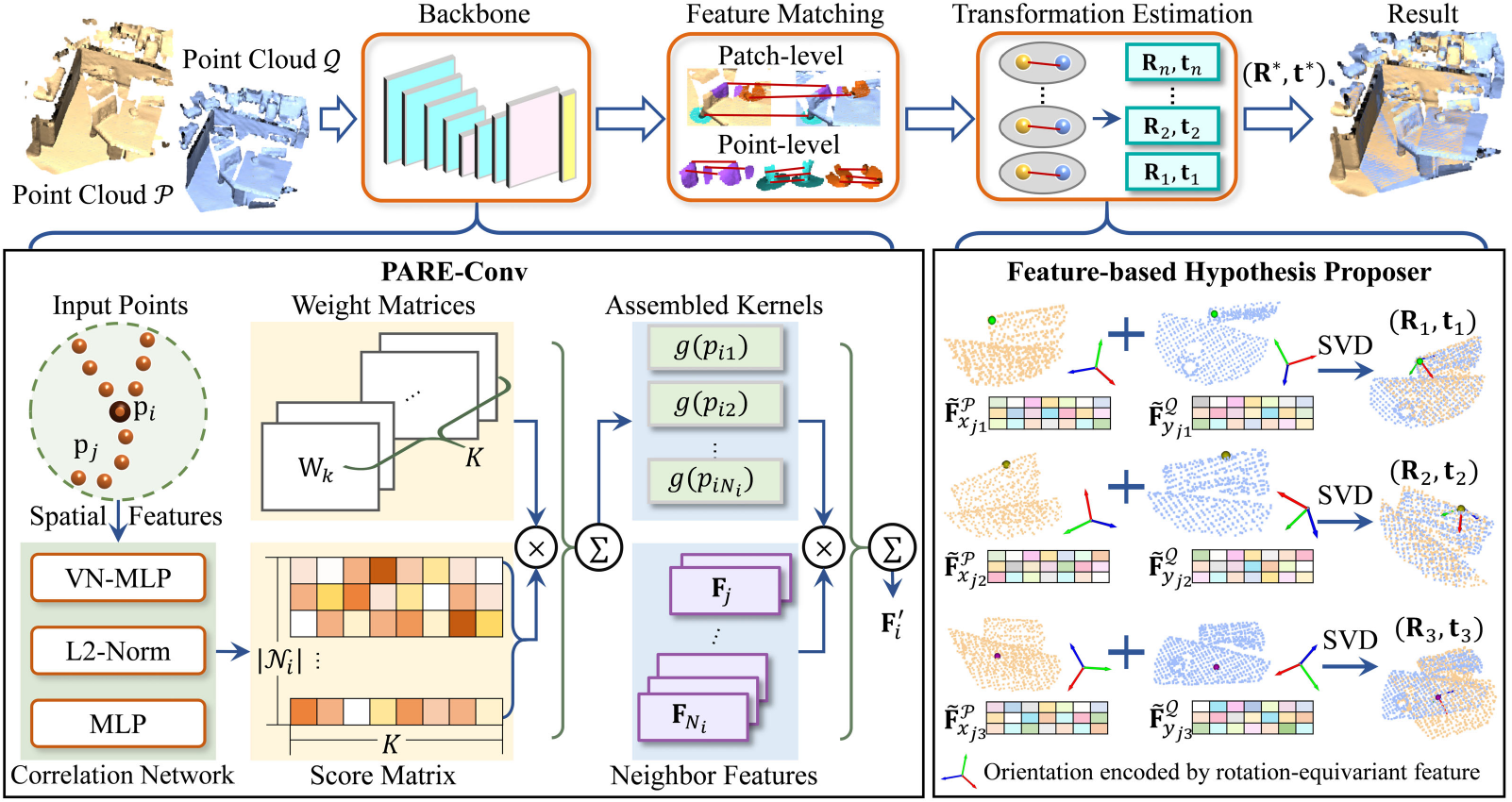}
    \caption{The framework of our method. Given point clouds $\mathcal{P}$ and $\mathcal{Q}$, a hierarchical backebone based on PARE-Conv is used to extract patch-level and point-level features. Then, we follow a coarse-to-fine scheme to obtain point correspondences using rotation-invariant features. Finally, the feature-based hypothesis proposer leverages the matched rotation-equivariant features ${\tilde{ \mathbf{F} }}^{\mathcal{P}}_{x_{jn}}$ and ${\tilde{ \mathbf{F}} }^{\mathcal{Q}}_{y_{jn}}$, which encode fine-grained structure orientations, to generate multiple reliable hypotheses. The best hypothesis is selected and refined as the final solution ($\mathbf{R}^*$, $\mathbf{t}^*$).  }
    \label{fig:framework}
\end{figure}
Given two partially overlapped point clouds $\mathcal{P} = \{\mathbf{p}_i \in \mathbb{R}^3 | i=1,2,\cdots, N\}$ and $\mathcal{Q} = \{\mathbf{q}_j \in \mathbb{R}^3 | j=1,2,\cdots, M\}$, we aim to estimate a transformation $\mathbf{T}=(\mathbf{R} \in SO(3), \mathbf{t} \in \mathbb{R}^3)$ that aligns the source point cloud to the target one. 
To address this problem, we propose a registration framework based on rotation-equivariant network $f(\mathbf{X})$, whose output is equivariant to the SO(3) action $\mathbf{R}$ applied on the input point cloud $\mathbf{X}$, i.e.,  $f(\mathbf{R} \circ \mathbf{X}) = \mathbf{R} \circ f(\mathbf{X})$. As we can see, rotation-equivariant features $f(\mathbf{R} \circ \mathbf{X})$ couple the structure information $f(\mathbf{X})$ and orientation information $\mathbf{R}$ of point clouds, which can be further decoupled for feature matching and transformation estimation, respectively. 

The framework of PARE-Net is depicted in Fig. \ref{fig:framework}. We build a hierarchical backbone based on PARE-Conv to extract both patch-level and point-level rotation invariant/equivariant features. Correspondences are estimated in a coarse-to-fine fashion by using rotation-invariant features. After that, the feature-based hypothesis proposer employs rotation-equivariant features to generate multiple reliable hypotheses, and the best one is selected as the final output ($\mathbf{R}^*$, $\mathbf{t}^*$).

\subsection{Position-Aware Rotation-Equivariant Network}

{\bf PARE-Conv}. Recent point cloud registration approaches \cite{predator, CoFiNet, qin2022geometric, peal} mainly exploited KPConv \cite{kpconv} to build backbones. However, KPConv is inherently rotation-sensitive and needs data augmentation to learn an approximate invariant mapping. As a result, models trained on limited samples of continuous SO(3) space are fragile to rotation variations. Moreover, enforcing the network to learn a rotation-invariant mapping inevitably requires more model parameters and also hinders the distinctiveness of features. Inspired by dynamic and conditional convolutions \cite{condconv, paconv}, we propose position-aware rotation-equivariant convolution to learn distinctive rotation-invariant/equivariant features, as shown in Fig. \ref{fig:framework}.

Given a point cloud $\mathcal{P} = \{\mathbf{p}_i \in \mathbb{R}^3 | i=1,2,\cdots, N\}$ and attached feature map $\mathbf{F} = \{\mathbf{F}_i \in \mathbb{R}^{C\times3}\}$, the general convolution of $\mathbf{F}$ by a kernel $g$ at a point $\mathbf{p}_i$ is defined as:
\begin{equation}
    (\mathbf{F} * g)(\mathbf{p}_i) = \sum_{\mathbf{p}_j \in \mathcal{N}_i} {g(\mathbf{p}_j - \mathbf{p}_i) \mathbf{F}_j},
\end{equation}
where $\mathcal{N}_i$ is the K-nearest neighbor set of the point $\mathbf{p}_i$. The core problem is the definition of kernel function $g$. In vanilla VN \cite{vn}, it omits the position information and simply uses a linear layer, $ \mathbf{W}{\mathbf{F}_i}$, followed by a VN non-linear layer and a VN pooling layer to aggregate local information, where $\mathbf{W} \in \mathbb{R}^{C'\times C}$  is a learnable weight matrix. Instead, we define a set of shadow kernel points, equipped with learnable weight matrices $\{\mathbf{W}_k|k=1,2,\cdots,K \}$. In image convolution, each pixel corresponds to a convolutional kernel, and the fixed spatial arrangement of the kernels enables the learning of the distribution information of image pixels. Due to the irregularity of point clouds, establishing such point-to-kernel correspondences is intractable. We follow KPConv \cite{kpconv} to establish soft assignments between data points and kernel points and the kernel function $g$ is defined as: 
\begin{equation}
    g(\mathbf{p}_{ij}) = \sum_{k=1}^{K}{\text{softmax}(\gamma(\mathbf{p}_{ij}, k))\mathbf{W}_k},
\end{equation}
where $\mathbf{p}_{ij}=\mathbf{p}_j - \mathbf{p}_i$ is the relative coordinate of $\mathbf{p}_{j}$ and $\gamma(\cdot)$ is a correlation function. 
In KPConv \cite{kpconv}, it uses a linear correlation $\gamma(\mathbf{p}_{ij}, k) = \text{max} (0, 1 - \|\mathbf{p}_{ij} - \tilde{\mathbf x}_k\| / \sigma)$, where $\tilde{\mathbf x}_k$ is the 3D coordinate of the $k$-th kernel point and $\sigma$ is a distance influence factor. However, kernel-to-point distances are sensitive to rotations thus cannot guarantee a rotation-equivariant convolution. Therefore, the key challenge lies in ensuring the rotation equivariance of the convolution while well capturing the spatial information of patches. Inspired by PAConv\cite{paconv}, we use a mini-network that consumes rotation-equivariant spatial features and predicts rotation-invariant correlation scores between data points and kernel points. Note that predicting rotation-invariant scores is necessary to guarantee the rotation-equivariance of the convolution (see Appendix A for the proof).  As shown in Fig. \ref{fig:framework}, we implement this function with 1) a VN-MLP block, learning geometric features $\mathbf{X}_{j} \in \mathbb{R}^{d\times 3}$ from rotation-equivariant spatial statistics, such as $\mathbf{p}_{ij}$, $\frac{1}{|\mathcal{N}_i|}\sum_{j}\mathbf{p}_{ij}$ and their cross product; 2) a L2-Norm layer, computing the magnitude of the vector neurons $\tilde{\mathbf{X}}_{j}=\|\mathbf{X}_{j}\|_2 \in \mathbb{R}^d$ to transform the feature into a rotation-invariant form; and 3) a MLP block, mapping this feature vector into K dimension to obtain the correlation vector, $\gamma_j=\text{MLP}(\tilde{\mathbf{X}}_{j}) \in \mathbb{R}^K$. Finally, a softmax operation is leveraged to output correlation scores. As a result, our convolution is formulated as:
\begin{equation}
    (\mathbf{F} * g)(\mathbf{p}_i) = \sum_{\mathbf{p}_j \in \mathcal{N}_i}\sum_{k} {\gamma_{jk}\mathbf{W}_k \mathbf{F}_j}.
\end{equation}
Compared to KPConv using a linear correlation, our correlation network can learn more complex relationships. Moreover,
PARE-Conv predicts correlation scores depending on local spatial property, thus can leverage the spatial information better and can provide stronger expressive power to extract more discriminative descriptors. Moreover, it is compatible with DGCNN \cite{wang2019dynamic}, allowing replacing the node feature $\mathbf{F}_j$ with the edge feature $[\mathbf{F}_j - \mathbf{F}_i, \mathbf{F}_j ]$. This can further enhance the expressive power, as discussed in the ablation study \ref{sec:ablation}.

{\bf Fully Convolutional Network}.
As shown in Fig. \ref{fig:framework}, we develop a hierarchical convolution network based on PARE-Conv. The original point clouds $\mathcal{P}$ and $\mathcal{Q}$ are downsampled three times to yield sparse superpoints $\mathcal{\hat{P}}$ and $\mathcal{\hat{Q}}$, whose resolution is $\frac{1}{2^3}$ of the original point clouds. Consequently, three convolution blocks are leveraged to extract multi-level features. The bottleneck rotation-equivariant features are denoted as $\mathbf{\hat{F}}^{\mathcal{P}} \in \mathbb{R}^{|\mathcal{\hat{P}}|\times \hat{d} \times 3} $ and $\mathbf{\hat{F}}^{\mathcal{Q}} \in \mathbb{R}^{|\mathcal{\hat{Q}}|\times \hat{d} \times 3}$, and a VN-invariant layer \cite{vn} is used to obtain the rotation-invariant features $\mathbf{\hat{X}}^{\mathcal{P}} \in \mathbb{R}^{|\mathcal{\hat{P}}|\times 3\hat{d}} $ and $\mathbf{\hat{X}}^{\mathcal{Q}} \in \mathbb{R}^{|\mathcal{\hat{Q}}|\times 3\hat{d}}$. For the decoder, two nearest upsampling blocks are used to get the rotation-equivariant features $\tilde{\mathbf{F}}^{\mathcal{P}} \in \mathbb{R}^{|{\tilde{\mathcal{P}}|\times \tilde{d} \times 3}}$ and $\tilde{\mathbf{F}}^{\mathcal{Q}} \in \mathbb{R}^{|{\tilde{\mathcal{Q}}|\times \tilde{d} \times 3}}$ of the first-level downsampled points $\tilde{\mathcal{P}}$ and $\tilde{\mathcal{Q}}$. Another VN-invariant layer is leveraged to get their rotation-invariant features $\tilde{\mathbf{X}}^{\mathcal{P}} \in \mathbb{R}^{|{\tilde{\mathcal{P}}|\times 3\tilde{d}}}$ and $\tilde{\mathbf{X}}^{\mathcal{Q}} \in \mathbb{R}^{|{\tilde{\mathcal{Q}}|\times 3\tilde{d}}}$. For more details on the convolutional block, please refer to Appendix B.

Following GeoTrans \cite{qin2022geometric}, we exploit the point-to-node grouping strategy to assign the points $\tilde{\mathcal{P}}$ to the superpoints $\mathcal{\hat{P}}$. The subset of $\tilde{\mathcal{P}}$ belongs to the superpoint $\mathcal{\hat{P}}_i$ is denoted as $\mathcal{G}^{\mathcal{P}}_i$ and the associated rotation-equivariant and invariant feature matrices of subset points are denoted as $\tilde{\mathbf{F}}^{\mathcal{P}}_i$ and $\tilde{\mathbf{X}}^{\mathcal{P}}_i$. For point cloud $\tilde{\mathcal{Q}}$, the points are grouped as $\{\mathcal{G}^{\mathcal{Q}}_i\}$, and the feature matrices are denoted as $\tilde{\mathbf{F}}^{\mathcal{Q}}_i$ and $\tilde{\mathbf{X}}^{\mathcal{Q}}_i$ in a same way.  

\subsection{Coarse-to-Fine Matching}
Since our backbone outputs both patch-level and point-level features, we use a coarse-to-fine matching strategy \cite{CoFiNet, qin2022geometric, roitr} to filter out non-overlapped regions and estimate more accurate correspondences.

\textbf{Superpoint Matching.} Direct matching superpoints by nearest neighbor search lacks robustness to repeated patterns and low-overlapped problems. Therefore, we utilize the Geometric Transformer module \cite{qin2022geometric} presented by Qin et al. to reason the global context of two point clouds, which iteratively uses self and cross-attention to capture intra and inter-point-cloud features.

We feed the rotation-invariant features $\hat{\mathbf{X}}^{\mathcal{P}}$ and $\hat {\mathbf{X}}^{\mathcal{Q}}$ and the attached position information $\mathcal{\hat{P}}$ and $\mathcal{\hat{Q}}$ into this module to get more discriminative features $\hat{\mathbf{H}}^{\mathcal{P}}$ and $\hat{\mathbf{H}}^{\mathcal{Q}}$. Then, a Gaussian correlation matrix is computed, followed by a dual normalization operation. The superpoint correspondences are obtained by selecting the top-\textit{k} reliable correspondences $\hat{\mathcal{C}}=\{({\hat{\mathbf{p}}}_{x_i}, {\hat{\mathbf{q}}}_{y_i}) | (x_i, y_i) \in \arg\max_{x, y} \mathbf{S}_{x,y}\}$, where $\mathbf{S}$ is the similarity matrix.

{\it Improvement}. Here, we find this module is sensitive to the overlap distribution of the dataset due to its cross-attention layers. This could lead to a significant performance decrease when there is a substantial difference between the overlap distributions of the test set and the training set. To tackle this issue, we present a simple but effective data augmentation technique, RandomCrop, where point clouds are randomly cropped to make the model more robust to low-overlapped point cloud pairs. We present more details in the Appendix C.

\textbf{Point Matching.}
When the correspondences $\hat{\mathcal{C}}_i=({\hat{\mathbf{p}}}_{x_i}, {\hat{\mathbf{q}}}_{y_i})$ of superpoints are established, we search point-level correspondences within $\mathcal{G}^{\mathcal{P}}_{x_i}$ and $\mathcal{G}^{\mathcal{Q}}_{y_i}$. We follow \cite{lightglue} that disentangles the similarity and saliency of features by using a matchability head $\mathbf{W}_m: \mathbb{R}^{3\tilde{d}} \rightarrow {\mathbb{R}}^{3\tilde{d}}$ and a saliency head $\mathbf{W}_s: \mathbb{R}^{3\tilde{d}} \rightarrow {\mathbb{R}}$. Specifically, we compute a matching matrix:
\begin{equation}
  \mathbf{M}_{x_i,y_i} = (\mathbf{W}_m\mathbf{X}^{\mathcal{P}}_{x_i})^{\text{T}} \mathbf{W}_m\mathbf{X}^{\mathcal{Q}}_{y_i}/ \sqrt{3\tilde{d}},
\end{equation}
and then the salient scores are predicted:
\begin{equation}
\sigma_{x_i}^{\mathcal{P}}=\text{Sigmoid}\left(\mathbf{W}_s\mathbf{X}^{\mathcal{P}}_{x_i}\right ),\sigma_{y_i}^{\mathcal{Q}}=\text{Sigmoid}\left(\mathbf{W}_s\mathbf{X}^{\mathcal{Q}}_{y_i} \right ).
\end{equation}
Finally, a soft assignment matrix $\mathbf{Z}_i$ is computed by:
\begin{equation}
    \mathbf{Z}_i = \sigma_{x_i}^{\mathcal{P}}\sigma_{y_i}^{\mathcal{Q}}\varphi(\mathbf{M}_{x_i,y_i})_{x_i}\varphi(\mathbf{M}_{x_i,y_i})_{y_i},
\end{equation}
where $\varphi$ is the softmax operation. As a result, we select $N_f$ point-level correspondences  $\mathcal{C}=\{(\tilde{\mathbf{p}}_{x_j}, \tilde{\mathbf{q}}_{y_j})\}$ across all patch correspondence whose assignment scores are the highest. The impact of $N_f$ is analyzed in Fig. \ref{fig:hypo}. 

\subsection{Feature-based hypothesis Proposer}
\begin{wrapfigure}{r}{0.35\textwidth}
    \centering
    \includegraphics[width=.35\textwidth]{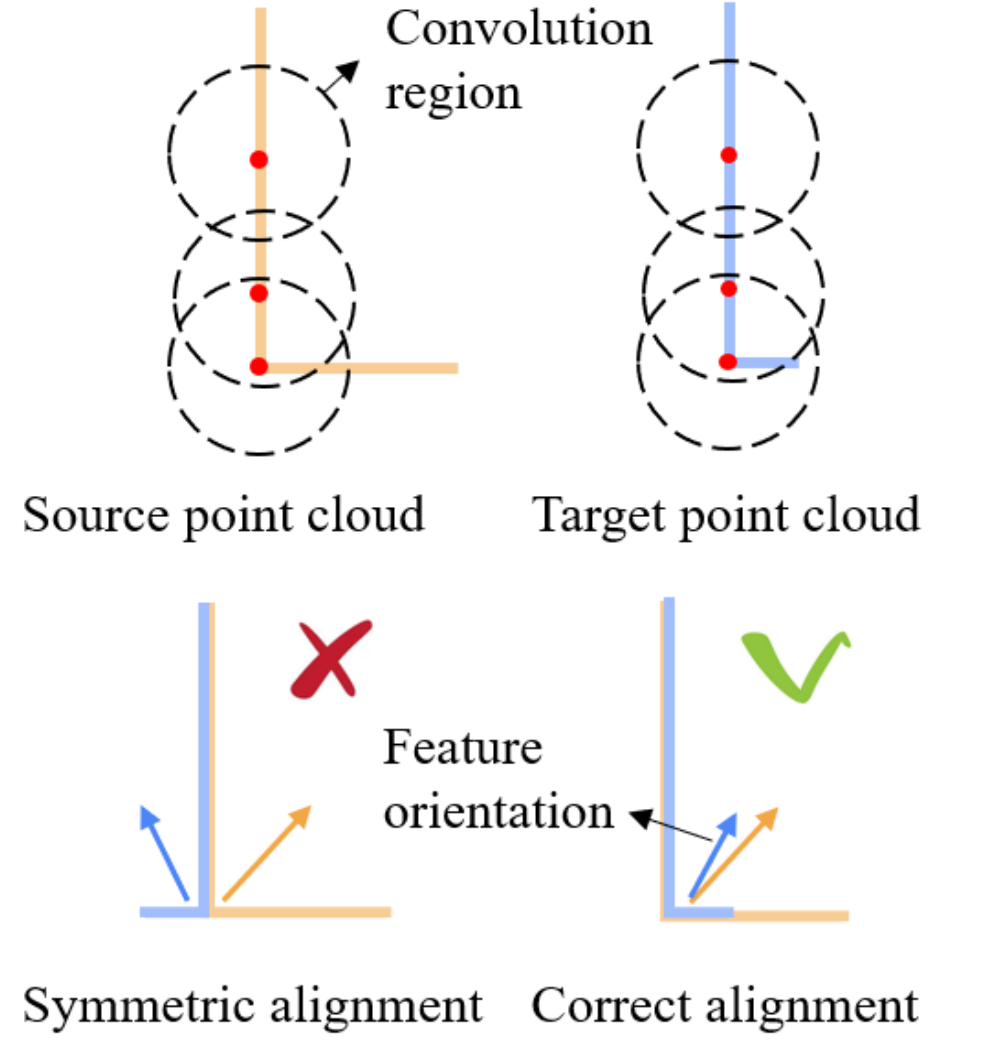}
    \caption{An illustration of incorrect alignment caused by the ambiguity of using coordinates to align. This problem can be solved by using rotation-equivariant features that contain fine-grained local orientation information. }
    \label{fig:ambiguity}
 \end{wrapfigure}

Since our rotation-equivariant features encode orientation information of local structures, we derive the pose transformation from matched features. 

\textbf{Hypothesis Generation and Selection.} Once obtaining point-wise correspondence set $\mathcal{C}=\{(\tilde{\mathbf{p}}_{x_j}, \tilde{\mathbf{q}}_{y_j})\}$, we leverage their corresponding rotation-equivariant feature maps $\{(\tilde{\mathbf{F}}^{\mathcal{P}}_{x_j}, \tilde{\mathbf{F}}^{\mathcal{Q}}_{y_j})\}$ to generate multiple model hypotheses. For each correspondence $({\tilde{\mathbf{p}}}_{x_j}, {\tilde{\mathbf{q}}}_{y_j})$, a rotation can be estimated by solving:
\begin{equation}
    \tilde{\mathbf{R}}_j = \mathop {\arg \min }\limits_{\mathbf{R}} \|\mathbf{R}(\tilde{\mathbf{F}}^{\mathcal{P}}_{x_j})^\text{T}- (\tilde{\mathbf{F}}^{\mathcal{Q}}_{y_j})^\text{T}\|_2^2.
\label{solve_R}
\end{equation}
This can be efficiently solved by SVD \cite{besl1992method} in closed form. Subsequently, the translation can be solved by:
\begin{equation}
    \tilde{\mathbf{t}}_j = {\tilde{\mathbf{q}}}_{y_j} - \tilde{\mathbf{R}}_j {\tilde{\mathbf{p}}}_{x_j}.
\label{solve_t}
\end{equation}
As a result, we can get a set of model hypotheses $\mathcal{H}_c=\{(\tilde{\mathbf{R}}_i, \tilde{\mathbf{t}}_i)\}$. The hypothesis that has the most inlier support will be selected as the alignment solution:
\begin{equation}
    (\mathbf{R}^*, \mathbf{t}^*) = \mathop {\arg\max}_{\tilde{\mathbf{R}}_j, \tilde{\mathbf{t}}_j} \sum_{(\tilde{\mathbf{p}}_{x_j}, \tilde{\mathbf{q}}_{y_j}) \in \mathcal{C}} \mathds{1}\left( \|{\tilde{\mathbf{R}}_j}\tilde{\mathbf{p}}_{x_j}+\tilde{\mathbf{t}}_j - \tilde{\mathbf{q}}_{y_j}\|^2_2 < \tau_d \right),
\end{equation}
where $\tau_d$ is the acceptance radius. The best model hypothesis will be iteratively refined by its inlier correspondences for a more accurate solution, as in \cite{qin2022geometric}.

\textbf{Discussion}. Traditional estimators, such as RANSAC \cite{ransac} randomly sample multiple pairs of correspondences and use their coordinates to estimate the transformation.  Only when all the sampled correspondences belong to inliers, a reliable transformation can be estimated. This approach is inefficient and requires multiple iterations. In contrast, we directly utilize rotation-equivariant features to estimate the pose of the point cloud, and each pair of correspondences generates a hypothesis. This approach is more efficient, as depicted in the Fig. \ref{fig:hypo}. Our method achieves a significantly high registration recall by generating only 1000 hypotheses. In contrast, RANSAC is still inferior to our method even with 50,000 iterations. Moreover, using coordinates to solve the transformation may falsely align the point clouds with opposite orientations, due to ignoring the orientation information, as shown in Fig. \ref{fig:ambiguity}. This problem can be easily addressed by our method as the rotation-equivariant features encode orientations of local structures. In addition, some methods \cite{roreg, buffer} use regressors to implicitly estimate transformations from equivariant features. Instead, our method can directly solve pose estimation through feature alignment, which is conceptually simpler and computationally efficient.

\subsection{Loss Function}
The network is supervised by three losses $\mathcal{L} = \mathcal{L}_c + \mathcal{L}_f+\mathcal{L}_r$, including superpoint matching loss $\mathcal{L}_c$, point matching loss $\mathcal{L}_f$, and contrastive rotation loss $\mathcal{L}_r$.

\textbf{Point Matching Loss $\mathcal{L}_f$}. Since point matching is performed within the scopes of matched superpoints, unreliable superpoint matching may lead to sparse supervision signals for point matching. Therefore, we use positive superpoint correspondences computed by ground-truth transformation during the training process. For two grouped point sets $\mathcal{G}^{\mathcal{P}}_{x_i}$ and $\mathcal{G}^{\mathcal{Q}}_{y_i}$, the positive correspondences $\mathcal{C}^*_i$, having distance errors smaller than a threshold $d_p$, are identified by the groundtruth. The unmatched points are denoted as ${\mathcal{I}}_{i}$, ${\mathcal{J}}_{i}$. We compute the log-likelihood loss on soft assignment matrix $\mathbf{Z}_i$ and the salient scores $\sigma_{x_i}$, $\sigma_{y_i}$:
\begin{equation}
    \begin{aligned} 
        \mathcal{L}_{f_i} = &-\frac{1}{|\mathcal{C}^*_i|}\sum_{(x_i, y_i) \in \mathcal{C}^*_i} {\log{\mathbf{Z}_{x_i, y_i}}} -\frac{1}{2|\mathcal{I}_i|}\sum_{x_i \in \mathcal{I}_i} {\log{(1-\sigma^{\mathcal{P}}_{x_i})}}\\&-\frac{1}{2|\mathcal{J}_i|}\sum_{y_i \in \mathcal{J}_i} {\log{(1-\sigma^{\mathcal{Q}}_{y_i})}}.
    \end{aligned}
\end{equation}
The overall point matching loss is a sum of $\mathcal{L}_f=\sum_{i}{\mathcal{L}_{f_i}}$.

\textbf{Contrastive Rotation Loss $\mathcal{L}_{r}$}. Rotation-equivariant feature inherently encodes pose information of local geometric structure. However, such features of a positive correspondence inevitably be contaminated by data degradation, such as occlusion, noise and down-sampling errors. To make the features more robust to these troubles, we propose a contrastive rotation loss, to supervise point-level correspondences:
\begin{equation}
    \begin{aligned}
       \mathcal{L}_{r_i}&= \frac{1}{|\mathcal{C}^*_i| \cdot \tilde d}\sum_{(x_i, y_i) \in \mathcal{C}^*_i} \sum_{c=1}^{\tilde d} [ \|{\tilde{\mathbf{F}}^{\mathcal{P}}_{x_i, c}\mathbf{R}^{\text{T}}_{gt} - \tilde{\mathbf{F}}^{\mathcal{Q}}_{y_i, c}}\|^2_2 - \alpha]_+\\&+  \frac{1}{|\mathcal{\bar C}_i| \cdot \tilde d}\sum_{(x_i, y_i) \in \mathcal{\bar C}_i} \sum_{c=1}^{\tilde d} [ \beta -\|{\tilde{\mathbf{F}}^{\mathcal{P}}_{x_i, c}\mathbf{R}^{\text{T}}_{gt} - \tilde{\mathbf{F}}^{\mathcal{Q}}_{y_i, c}}\|^2_2 ]_+,
    \end{aligned}
\end{equation}
where $\mathcal{C}^*_i$ is the aforementioned positive correspondence set, $\mathcal{\bar C}_i$ is the negative correspondence set whose point pair has spatial distance greater than $d_n$, $\alpha=0.1$ and $\beta=1.4$ are positive and negative margins, and  $[\cdot]_+$ is the ReLU operation.

\textbf{Superpoint Matching Loss $\mathcal{L}_c$}. To supervise the rotation-invariant features of superpoints, we utilize the overlap-aware circle loss that considers the overlap ratio of two superpoints to compute more accurate gradients. For more details, please refer to GeoTrans \cite{qin2022geometric}.

\section{Experiments}
To evaluate the performance of our method, we compare our method with state-of-the-art approaches and perform extensive experiments on 3DMatch \cite{zeng20173dmatch}, 3DLoMatch \cite{predator}, and KITTI Odometry \cite{geiger2012we}. Finally, ablation studies are conducted to analyze the components of our method. All the experiments are conducted with a RTX 3090 GPU with Intel (R) Xeon (R) Silver 4314 CPU. More implementation details are reported in Appendix D. 

\subsection{Indoor Dataset: 3DMatch \& 3DLoMatch}
\textbf{DataSet}. 3DMatch \cite{zeng20173dmatch} and 3DLoMatch \cite{predator} are collected from indoor RGB-D datasets. 3DMatch contains 62 scenes and there are 46/8/8 scenes for training/validation/testing. It consists of point cloud pairs with overlap ratios greater than 30\%. 3DLoMatch is only used for testing that contains point cloud pairs with overlap ratios from 10\% to 30\%. We follow the protocols \cite{predator,roreg, qin2022geometric} to perform all the experiments.

\textbf{Baseline}.
We compare our method with strong baselines: 1) FCGF \cite{fcgf}, Predator \cite{predator}, GeoTrans \cite{qin2022geometric}, and PEAL \cite{peal} are advanced scene-level feature extractors, and some of them \cite{qin2022geometric, peal} follow a coarse-to-fine manner to extract robust correspondences. PEAL requires overlap prior as input, we follow the office implementation that leverages 3D overlap prior predicted by GeoTrans for a fair comparison. 2) YOHO \cite{yoho}, RoReg \cite{roreg}, and RoITR \cite{roitr} are also scene-level feature extractors but focus on rotation-invariant/equivariant network designs. 3) SpinNet \cite{spinnet} and BUFFER \cite{buffer} leverage patch-level feature extractors with elaborate designs for rotation-invariance.

\textbf{Metrics}.
We follow \cite{predator, qin2022geometric} that use three evaluation metrics, including Inlier Ratio (IR), Feature Matching Recall (FMR), and Registration Recall (RR). To evaluate the performance of different methods on rotated datasets, RoITR \cite{roitr} uses two ways to compute the RMSE on the original and rotated datasets, which cannot show the robustness of methods against pose variations. To tackle this issue,  we follow \cite{roreg} to report Rotation Error (RE), Translation Error (TE), and Transformation Recall (TR) on both original and rotated datasets. Please refer to the Appendix E for more details about these evaluation metrics.

\begin{table}[tbp]
  \centering
  \caption{Evaluation results on 3DMatch and 3DLoMatch. Methods with the RANSAC estimator are marked by $\diamond$, which exploit 5000 points to establish correspondences. }
  \label{tab:Indoor}
  \tabcolsep=0.3cm
  \resizebox{\linewidth}{!}{
  \begin{tabular}{l|r|cccc|cccc}
    \toprule
    \multirow{2}{*}{Method} & \multirow{2}{*}{\shortstack{Size \\(MB)}} & \multicolumn{4}{c|}{3DMatch} & \multicolumn{4}{c}{3DLoMatch} \\
    \cline{3-10}
     \rule{0pt}{19pt}&  & {\shortstack {FMR \\(\% $\uparrow$)}}  &{\shortstack{IR \\(\% $\uparrow$)}}    &{\shortstack{RR \\(\% $\uparrow$)}}   & {\shortstack{Time \\(s $\downarrow$)}} & {\shortstack{FMR \\(\% $\uparrow$)}}  & {\shortstack{IR \\(\% $\uparrow$)}} &{\shortstack{RR \\(\% $\uparrow$)}}   & {\shortstack{Time \\(s $\downarrow$)}} \\
    \hline
    FCGF$^\diamond$ \cite{fcgf} & 8.76 & 94.7 & 31.1 & 82.8 & \textbf{0.12}   & 59.4 & 9.8 & 38.0 & \textbf{0.13}\\
    SpinNet$^\diamond$ \cite{spinnet} & \underline{1.41} & 97.6 & 47.5 & 88.6 & 9.85  & 75.3 & 20.5 & 59.8 &9.03 \\
    YOHO \cite{yoho} & 12.38 & 98.2 & 64.4 & 90.8   & 2.81 & 78.9 & 25.9 & 66.0 & 2.62 \\
    Predator$^\diamond$ \cite{predator} & 7.43 & 96.6 & 58.0 & 89.0   &0.64  & 78.2 & 26.7 & 64.4 & 0.47  \\
    GeoTrans \cite{qin2022geometric} & 9.83 & 98.1 & 70.9 & 92.4  &0.18   & 87.4 & 43.5 & 74.3 &\underline{0.17}\\
    RoReg \cite{roreg} & 12.71 & 98.2 & \underline{81.6} & 93.0 &2.27 & 82.3 & 39.6 & 70.1 & 2.10 \\
    BUFFER \cite{buffer} & \textbf{0.92 }& - & - & 92.1  &0.22   & - &- & 70.3 &0.21\\
    RoITR$^\diamond$\cite{roitr} &10.10 & 98.0 & \textbf{82.4} & 91.9  & 0.36  & \textbf{89.2} & \textbf{54.6} & 74.1 &0.34 \\
    PEAL\cite{peal} &9.83 &\underline{98.4} & 71.0 & \underline{94.2}  &1.46   & \underline{88.3} & 46.0 & 78.8 &1.19 \\
    Ours & 3.84 & \textbf{98.5} & 76.9 & \textbf{95.0} & \underline{0.17}  & \underline{88.3} & \underline{47.5} & \textbf{80.5} & \underline{0.17} \\
    \bottomrule
\end{tabular}}
\end{table}

\textbf{Quantitative results}.
We report the experimental results in Table \ref{tab:Indoor}. Compared with SOTA methods, our approach achieves a significantly high RR on both 3DMatch and 3DLoMatch, while simultaneously maintaining lightweight and computation efficient. PEAL \cite{peal} is a formidable competitor that employs an iterative manner for alignment due to the requirements of overlap prior, resulting in significant time overhead. In contrast, our method achieves a speed improvement of 7 times compared to PEAL, while also outperforming it by 1.7\% in terms of RR on 3DLoMatch. GeoTrans \cite{qin2022geometric}, RoITR \cite{roitr}, and our method all employ a coarse-to-fine matching framework. Despite using a more lightweight feature extractor, our RR is significantly higher than theirs, particularly on 3DLoMatch, where our method outperforms them by 6.2\% and 6.4\% respectively. This can be attributed to the ability of our rotation-equivariant network to focus more on learning discriminative structural information. YOHO \cite{yoho} and RoReg \cite{roreg} are extremely time-consuming due to the group feature extractor and embedder. Our efficiency advantage stems from two aspects. Firstly, we utilize a lightweight scene-wise feature extractor, which only needs a forward process to extract dense rotation-equivariant features. Secondly, our feature-based hypothesis proposer can simultaneously generate multiple valid hypotheses by using matched rotation-equivariant features. Thus, it is more efficient compared to RANSAC \cite{ransac} with many approaches \cite{fcgf, spinnet, predator, roitr} used.

\begin{table}[htbp]
  \centering
  \caption{Evaluation results on 3DLoMatch and Rotated 3DLoMatch. For easy comparison of the impact of rotation, the numerical changes w.r.t. TR on the rotated dataset are annotated in the top right corner of each value. Methods with rotation invariant/equivariant designs are marked by $*$ symbol. }
  \label{tab:rotation}
  \tabcolsep=0.15cm
    \resizebox{\linewidth}{!}{
  \begin{tabular}{l|r|ccc|ccl}
    \toprule
    \multirow{2}{*}{Method} & \multirow{2}{*}{\shortstack{Size \\(MB)}} & \multicolumn{3}{c|}{3DLoMatch} & \multicolumn{3}{c}{Rotated 3DLoMatch} \\
     && RE ($^\circ$ $\downarrow$) & TE (cm $\downarrow$)  &TR (\% $\uparrow$) & RE ($^\circ$ $\downarrow$) & TE (cm $\downarrow$)  &TR (\% $\uparrow$) \\
    \hline
    FCGF \cite{fcgf} & 8.76 &4.84&12.87&39.6& 4.74 & 13.39 & 24.5$^{-15.1}$\\
    PREDATOR \cite{predator} & 7.43&3.61 & 10.65 &65.6&3.55&10.30&64.0$^{-1.6}$ \\
    GeoTrans \cite{qin2022geometric} & 9.83 & 2.91 & \underline{8.71} & 75.4 & 2.94& 8.85 & 72.6 $^{-2.8}$ \\  
    PEAL \cite{peal} &9.83 &\textbf{2.84}&\textbf{8.64}&\underline{81.2}&\underline{2.86}&\textbf{8.53} &\underline{78.7}$^{-2.5}$\\
    \hline
    YOHO$^*$ \cite{yoho} & 12.38 &3.54&10.34&66.6&3.61&10.16&67.1$^{+0.5}$\\
    RoReg$^*$ \cite{roreg} & 12.71 & 3.01&9.26&71.3& 3.03 & 9.28&71.0$^{-0.3}$ \\
    BUFFER$^*$ \cite{buffer} & \textbf{0.92} &3.03&9.86&74.4&3.02&9.99&74.7$^{+0.3}$ \\
     RoITR$^*$ \cite{roitr} & 10.10 &2.95 & 9.03 & 75.1&2.97&9.08&75.5$^{+0.4}$ \\
    Ours$^*$ & \underline{3.84} &\underline{2.87} & 8.83 & \textbf{81.3}& \textbf{2.84} & \underline{8.71} & \textbf{81.8} $^{+0.5}$\\
    \bottomrule
  \end{tabular}}
\end{table}
\textbf{Experiments on the rotated dataset}. To test the robustness against rotations, we follow previous works \cite{roreg, roitr} that evaluate all the methods on the rotated 3DLoMatch, where arbitrary rotations are applied to point clouds. The experimental results are shown in Table \ref{tab:rotation}.  As we can see, PEAL experiences a decrease in TR by 2.5\% on the Rotated 3DLoMatch dataset, while our method demonstrates an increase of 0.5\%. This result validates the robustness of our approach against rotations, illustrating its ability to handle such variations. Methods without rotation-invariant design, such as FCGF \cite{fcgf}, Predator \cite{predator}, and GeoTrans \cite{qin2022geometric} also show performance degradation, indicating that networks trained with rotation augmentation are vulnerable to rotation variations. On the other hand, methods such as YOHO \cite{yoho}, RoReg \cite{roreg}, RoITR \cite{roitr}, and BUFFER \cite{buffer}, which incorporate rotation equivariant or invariant designs, demonstrate better robustness against data rotations. However, our method significantly outperforms these methods by 6\% w.r.t TR on both 3DLoMatch and Rotated 3DLoMatch. This is because our backbone can capture spatial geometric information better, thus extracting more distinctive descriptors.

\begin{wraptable}{r}{0.6\textwidth}
\centering
  \caption{Evaluation results on KITTI Odometry.}
  \label{tab:kitti} 
    \resizebox{\linewidth}{!}{ 
  \begin{tabular}{l|r|cccr}
    \toprule
    Method & Size & RE ($^\circ$) & TE (cm)  &TR (\%) & Time (s)\\
    \hline
    FCGF \cite{fcgf}& 8.76 & 0.30 & 9.5 &96.6&-\\
    D3Feat \cite{d3feat}& 14.08 & 0.30 & 7.2 & \textbf{99.8}&-\\
    Predator \cite{predator}&22.77 & 0.27 & 6.8 & \textbf{99.8}&0.77\\
    SpinNet \cite{spinnet} &1.41&0.47&9.9 & 99.1& 16.24\\
    CoFiNet \cite{CoFiNet}&5.48 &0.41 & 8.2 & \textbf{99.8}&0.59\\
    GeoTrans \cite{qin2022geometric} &25.50&\textbf{0.23}& 6.2 & \textbf{99.8} & 0.26\\
    BUFFER \cite{buffer} &\textbf{0.92}&0.26& 7.1 & \textbf{99.8} & 0.27\\
    Ours & 2.08 & \textbf{0.23} & \textbf{4.9} & \textbf{99.8} & \textbf{0.21} \\
    \bottomrule
  \end{tabular}}
\end{wraptable}

\begin{figure*}
    \centering
    \includegraphics[width=1\textwidth]{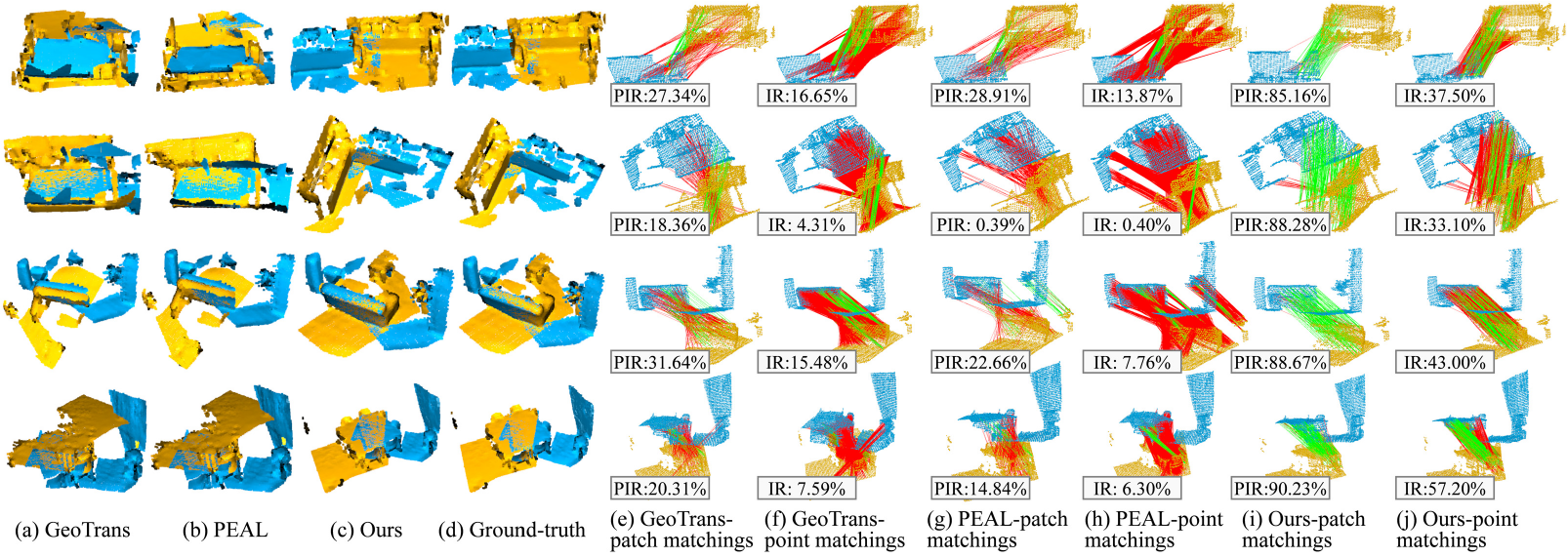}
    \caption{ Qualitative results. Our method can successfully align low-overlapped pairs with higher IR and PIR (Patch IR). Moreover, GeoTrans \cite{qin2022geometric} and PEAL \cite{peal} incorrectly align point clouds with opposite orientations (the first and second rows), our method successfully addresses this issue because the equivariant features contain orientation information of local structure, alleviating this ambiguity. }
    \label{fig:visua}
\end{figure*}

\textbf{Qualitative results}. We show some registration results in Fig. \ref{fig:visua}, which demonstrate that our method can estimate more accurate patch-level and point-level correspondences and can successfully align point cloud pairs with low-overlapped ratios. Moreover, GeoTrans \cite{qin2022geometric} and PEAL \cite{peal} incorrectly align point clouds with opposite orientations (the first and second rows), our method successfully addresses this issue because the equivariant features contain orientation information of local structure, alleviating this ambiguity. Please refer to Appendix G for more visualization results.

\subsection{Outdoor Dataset: KITTI Odometry}

\textbf{Dataset}. We follow previous methods \cite{fcgf, qin2022geometric} to evaluate our method on KITTI Odometry \cite{geiger2012we}. It comprises 11 urban scene sequences captured by a LiDAR sensor. We use 0-5/6-7/8-10 sequences for training/validation/testing, respectively. 

\textbf{Metrics}. We use the Rotation Error (RE), Translation Error (TE), and Transformation Recall (TR) to evaluate our method. Following \cite{fcgf, qin2022geometric}, the TR is computed with rotation threshold 5$^\circ$ and translation threshold 2m.

\textbf{Experimental results}. The experimental results are shown in Table \ref{tab:kitti}. On KITTI dataset, many methods have reached performance saturation, while our approach achieves the highest accuracy when achieving the highest TR. Our method also shows advantages w.r.t. model size and running speed.

\subsection{Ablation Study}
\label{sec:ablation}
To better understand the designs of our method, we conduct comprehensive ablation experiments and the evaluated results are reported in Table \ref{tab:ablation}.

\textbf{Ablation of backbone}. To investigate the impact of different convolution methods, we compare our PARE-Conv with VN \cite{vn}. We leverage both node features \cite{qi2017pointnet} and edge features \cite{wang2019dynamic} for evaluation.  The experimental results indicate that our approach achieves significant improvement compared to VN whether using node features or edge features. This demonstrates the enhanced representative power of PARE-Conv to learn more distinctive features. We also compare PARE-Conv with KPConv \cite{kpconv}, refer to the Appendix  F.5 for details.

\textbf{Ablation of pose estimator}.  To demonstrate the superiority of our pose estimator, we compared it with RANSAC \cite{ransac} and LGR \cite{qin2022geometric}. RANSAC randomly selects a few pairs of correspondences to generate hypotheses, while LGR generates a hypothesis using the correspondences within a matched patch pair. In contrast, our approach directly generates a hypothesis using the equivariant features of a pair of matched points. As shown in Table \ref{tab:ablation}, our method achieves the highest RR, outperforming RANSAC and LGR by 3\% and 1.6\% respectively on 3DLoMatch. This validates that estimating poses using rotation-equivariant features based on local structures is more robust than using point-to-point correspondences. Additionally, the results of FMR and IR are identical as we used the same correspondences for different pose estimators.

\begin{table}[htbp]
  \centering
  \caption{Ablation experiments of our method.} 
  \label{tab:ablation}
  \tabcolsep=0.04cm
    \resizebox{1\linewidth}{!}{
  \begin{tabular}{ll|ccl|ccl}
    \toprule
    \multirow{2}{*}{Component} &\multirow{2}{*}{Method} & \multicolumn{3}{c|}{3DMatch} & \multicolumn{3}{c}{3DLoMatch} \\
    \cline{3-8}
     & &FMR (\% $\uparrow$) & IR (\% $\uparrow$)  &RR (\% $\uparrow$) & FMR  (\% $\uparrow$)& IR (\% $\uparrow$)  &RR (\% $\uparrow$) \\
    \hline
    \multirow{4}{*}{Backbone} &VN (Node) \cite{vn} & 98.6 & 71.0 & 93.4 &87.8 & 41.4&76.6 \\
    &PARE-Conv (Node) & \textbf{98.9} & 74.1 & 94.5 $^{+1.1}$  & 88.3 & 44.0 & 78.8 $^{+2.2}$  \\
    &VN (Edge) \cite{vn}&98.7&72.4&94.1&\textbf{88.8}&43.5&77.8\\
    &PARE-Conv (Edge) & 98.5 & \textbf{76.9} & \textbf{95.0} $^{+0.9}$  & 88.1 & \textbf{47.5} & \textbf{80.5}  $^{+2.7}$  \\
    \hline
    \multirow{3}{*}{\shortstack{Pose \\ Estimator}} & RANSAC &\textbf{98.5}&\textbf{76.9}& 93.7& \textbf{88.3} & \textbf{47.5} &77.5 \\
    &LGR \cite{qin2022geometric} &\textbf{98.5}&\textbf{76.9}&94.0$^{+0.3}$&\textbf{88.3}&\textbf{47.5}&78.9 $^{+1.4}$\\
    &Ours & \textbf{98.5} & \textbf{76.9} & \textbf{95.0}$^{+1.3}$ & \textbf{88.3} & \textbf{47.5} & \textbf{80.5} $^{+3.0}$  \\
    \hline
    \multirow{3}{*}{\shortstack{Rotation \\ Loss}} & Without &98.3&74.4& 93.3 &\textbf{88.7} &45.5 & 79.3\\
    &L2 & \textbf{98.6} & 65.4 &94.2 $^{+0.9}$ & 85.9 & 35.0 &75.4$^{-3.9}$ \\
    &Contrastive & 98.5 & \textbf{76.9} & \textbf{95.0}$^{+1.7}$   & 88.3 & \textbf{47.5} & \textbf{80.5}$^{+1.2}$   \\
    \bottomrule
\end{tabular}}
\end{table}

\textbf{Ablation of rotation loss}. The contrastive rotation loss is designed to improve the robustness of rotation-equivariant features to data occlusion, noise, and partial overlap. When the rotation-equivariant features are not supervised by a loss function (Indeed, it is indirectly supervised by the loss function of rotation-invariant features, as the invariant features are derived from the equivariant features), the method shows decent performance on 3DLoMatch but shows poor performance on 3DMatch. When using the L2 loss to supervise equivariant features of positive point pairs, there is a slight improvement in RR on 3DMatch but shows performance degradation on 3DLoMatch. However, when utilizing our contrastive rotation loss, the performance of the method is enhanced on both 3DMatch and 3DLoMatch, demonstrating its superiority.




\section{Conclusion}
In this paper, we propose PARE-Net, a lightweight network for fast and robust point cloud registration. We introduce position-aware rotation-equivariant convolution to learn spatial information effectively, thus extracting more distinctive descriptors. We also propose an efficient feature-based hypothesis proposer to generate reliable model hypotheses. Extensive experiments demonstrate that our method significantly outperforms state-of-the-art approaches and shows great robustness to rotation variations. The limitations of our method and future works are discussed in Appendix H.

\textbf{Acknowledgements}.
This work was supported by the National Science and Technology Major Project of China under Grant No. 2020AAA0108102, the National Natural Science Foundation of China under Grant Nos. 62327808, 62073257 and 62088102, and the Fundamental Research Funds for the Central Universities under Grant No. xzy022024007.

\bibliographystyle{splncs04}
\bibliography{main}

\newpage
\title{PARE-Net: Position-Aware Rotation-Equivariant Networks for Robust Point Cloud Registration} 
\author{Supplementary Material}
\authorrunning{R. Yao et al.}
\institute{}
\maketitle
\renewcommand\thesection{\Alph{section}}
\setcounter{section}{0}
\setcounter{table}{4}
\setcounter{figure}{5}

This appendix provides theoretical proofs of the rotation-equivariant property of our PARE-Conv (Sec. \ref{proof}), details of network architecture (Sec. \ref{network}), detailed introduction of RandomCrop (Sec. \ref{augmentation}), implementation of our method (Sec. \ref{implentation}), and evaluation metrics (Sec. \ref{metrics}). We also report more experimental results, including quantitative results (Sec. \ref{quan_results}) and qualitative results (Sec. \ref{qual_results}). Finally, we discuss the limitation of our method and future work (Sec. \ref{limitation}).
\section{Theoretical Proof}
\label{proof}
Here, we theoretically demonstrate that our PARE-Conv is a rotation-equivariant mapping.

\textbf{Lemma 1}. \textit{The linear operation $f_{lin}(\mathbf{F})=\mathbf{W}\mathbf{F}$ is equivariant to rotations, where $\mathbf{F} \in \mathbb{R}^{C \times 3}$ is a vector-list feature and $\mathbf{W} \in \mathbb{R}^{C^{'} \times C}$ is a linear project matrix. }

\textbf{Proof}: Given a rotation $\mathbf{R} \in \text{SO(3)}$, we have:
\begin{equation}
    f_{lin}(\mathbf{F})\mathbf{R}=(\mathbf{W}\mathbf{F})\mathbf{R} = \mathbf{W}(\mathbf{F}\mathbf{R}) = f_{lin}(\mathbf{F}\mathbf{R}).
\end{equation}
Therefore, this linear operation is rotation-equivariant.

\textbf{Lemma 2}. \textit{The channel-wise concatenation operation $f_{cat}(\mathbf{F}_1, \mathbf{F}_2)=[\mathbf{F}_1 \|  \mathbf{F}_2] \in \mathbb{R}^{(C + C^{'}) \times 3}$  is equivariant to rotations, where $\mathbf{F}_1 \in \mathbb{R}^{C \times 3}$ and $\mathbf{F}_2 \in \mathbb{R}^{C^{'} \times 3}$ are two vector-list features. }

\textbf{Proof}: Given a rotation $\mathbf{R} \in \text{SO(3)}$, we have:
\begin{equation}
\begin{aligned}
    f_{cat}(\mathbf{F}_1, \mathbf{F}_2)\mathbf{R}&=[\mathbf{F}_1 \|  \mathbf{F}_2]\mathbf{R} = [\mathbf{F}_1\mathbf{R} \|  \mathbf{F}_2\mathbf{R}] \\&= f_{cat}(\mathbf{F}_1\mathbf{R}, \mathbf{F}_2\mathbf{R}).
\end{aligned}
\end{equation}
Therefore, the concatenation operation is rotation-equivariant.

\textbf{Lemma 3}. \textit{Linear combinations  
 $f_{lc} (f_1(\mathbf{F}),  f_2(\mathbf{F})) = \lambda_1f_1(\mathbf{F}) + \lambda_2f_2(\mathbf{F})$  of rotation-equivariant functions are still equivariant to rotations, where $\lambda_1$, $\lambda_2$ are coefficients and $f_1(\cdot)$, $f_2(\cdot)$ are rotation-equivariant functions.}

\textbf{Proof}: Given a rotation $\mathbf{R} \in \text{SO(3)}$, we have:
\begin{equation}
    \begin{aligned}
          f_{lc}(f_1(\mathbf{F}), f_2(\mathbf{F}))\mathbf{R}&=[\lambda_1f_1(\mathbf{F}) + \lambda_2f_2(\mathbf{F})]\mathbf{R}\\&= \lambda_1f_1(\mathbf{F})\mathbf{R} + \lambda_2f_2(\mathbf{F})\mathbf{R} \\&= \lambda_1f_1(\mathbf{F} \mathbf{R}) + \lambda_2f_2(\mathbf{F}\mathbf{R}) \\&= f_{lc}(f_1(\mathbf{F}\mathbf{R}), f_2(\mathbf{F}\mathbf{R})).
    \end{aligned}
\end{equation}
Therefore, the linear combinations of rotation-equivariant functions are still rotation-equivariant.

\textbf{Lemma 4}. \textit{The PARE-Conv is equivariant to rotations.}

\textbf{Proof}: We formulate PARE-Conv as $\sum\limits_{\mathbf{p}_j \in \mathcal{N}_i}\sum\limits_{k} {\gamma_{jk}\mathbf{W}_k \mathbf{F}_j}$. Since the KNN search and $\gamma_{jk}$ are both rotation-invariant, the PARE-Conv is the linear combination of $\mathbf{W}_k \mathbf{F}_j$. According to  \textbf{Lemma 1} and \textbf{Lemma 3}, our PARE-Conv achieves a rotation-equivariant manner. Moreover, if we replace the node feature $\mathbf{F}_j$ with edge feature $[\mathbf{F}_j - \mathbf{F}_i \| \mathbf{F}_j]$, it also satisfies rotation-equivariant according to \textbf{Lemma 2}. 

\section{Detailed Network Architecture}
\label{network}

We present some details of our network architecture.

\begin{figure}[htbp]
    \centering
    \includegraphics[width=0.5\textwidth]{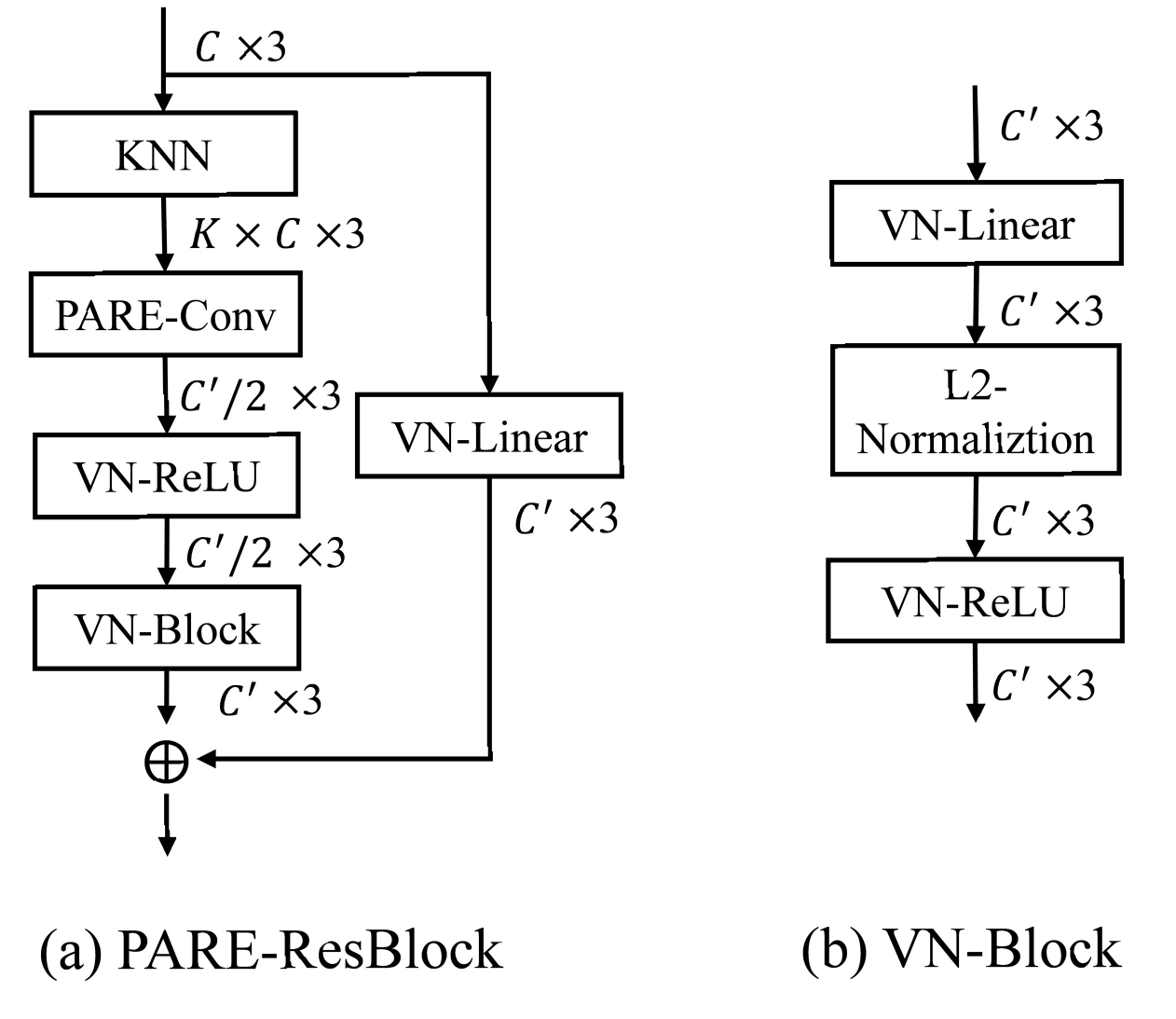}
    \caption{Illustration of PARE-ResBlock. $C$ and $C^{'}$ are the dimensions of the input and output features, respectively. The VN-Linear in the shortcut is only needed when $C \neq C^{'}$.}
    \label{fig:prresblock}
\end{figure}

\textbf{PARE-ResBlock}. Based on the PARE-Conv, we design a bottleneck block like ResNet \cite{resnet}, as shown in Fig. \ref{fig:prresblock} (a). We first use a PARE-Conv to learn local spatial features and squeeze the feature dimension into $C^{'} / 2$, followed by a VN-ReLU layer \cite{vn}. Then, a VN-block is leveraged to expand the dimension of features into $C^{'}$. Here, a shortcut is used to add the input to the output of the VN-block. Moreover, we detail the VN-block in Fig. \ref{fig:prresblock} (b), which consists of a VN-Linear, L2-Normalization, and VN-ReLU. In VN \cite{vn}, it presents a normalization layer by applying batch normalization to the magnitudes of vector-list features. However, we find it may harm the network convergence because the directions of the vectors may be flipped. Therefore, we only normalize the magnitudes of vectors into the unit length, which is so-called L2-Normalization. We find this modification facilitates the convergence of networks.

\begin{wrapfigure}{r}{0.35\textwidth}
    \centering
    \includegraphics[width=0.35\textwidth]{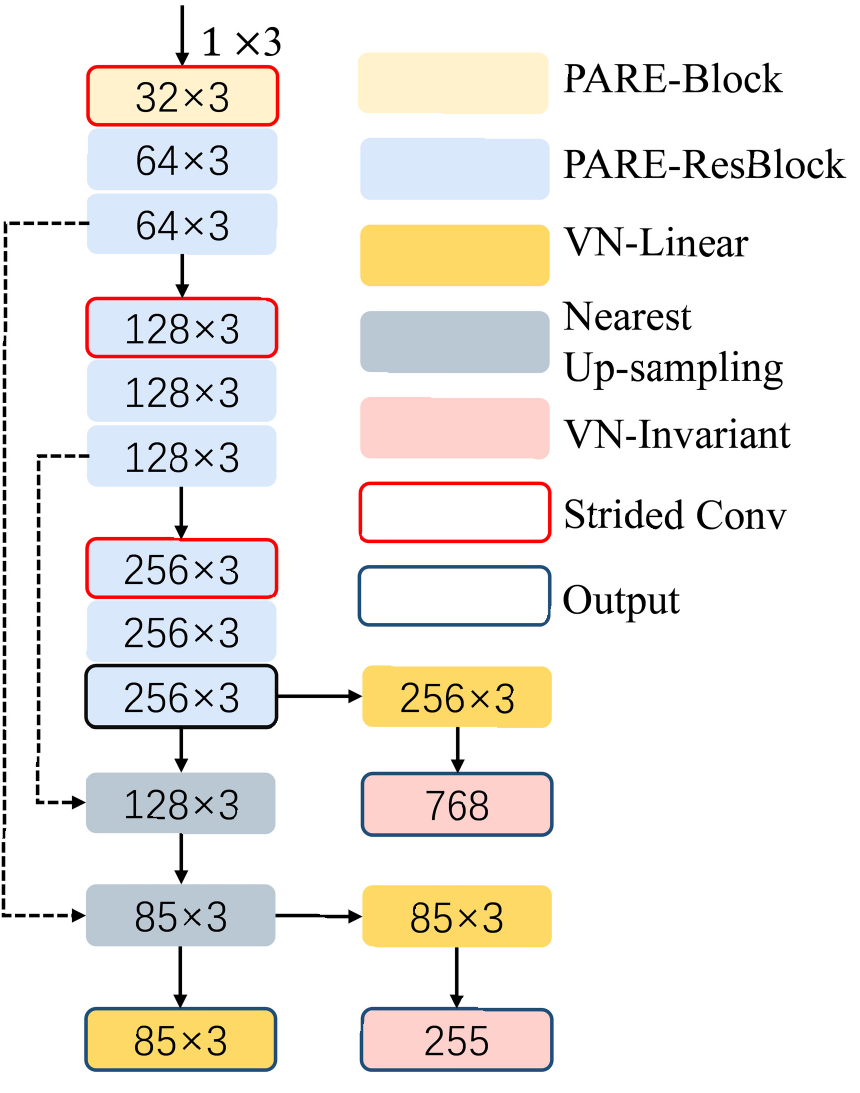}
    \caption{The detailed backbone of our method. }
    \label{fig:backbone} 
\end{wrapfigure}

\textbf{Backbone}. We build a hierarchical backbone to extract multi-level features, as shown in Fig. \ref{fig:backbone}. Different from works\cite{qin2022geometric, peal} that leverage four convolutional layers, we first downsample the input point clouds and then use three convolutional layers to learn features on sparse point clouds. This can make the network lighter and computation cheaper. Each layer contains three PARE-ResBlocks and the first one is the strided block that performs convolution on down-sampled points. After that, two nearest up-sampling layers are used to decode the features from sparser points to denser points. Specifically, skip links are leveraged to pass the intermediate features from the encoder to the decoder. For a point in the denser layer, its skipped feature is concatenated to the feature of the sparse point which is nearest to the dense point in Euclidean space, and these features are fused by a VN-Block.

Finally, we use two rotation-invariant layers to obtain the rotation-invariant features for points and superpoints. We share the same backbone for 3DMatch and KITTI. Here, a difference is that the dimension of point features is set to 255 for 3DMatch, while it is set to 63 for KITTI. However, the large-scale KITTI dataset may generate too many superpoints after three down-sampling operations. To address this issue, GeoTrans \cite{qin2022geometric} down-samples the point clouds four times and performs convolutions in five stages. This makes its network too heavy with 25.5 MB parameters. Instead, we increase the down-sampling ratio from 2 to 2.5 to make the supurpoints more sparse for the KITTI dataset. Therefore, we can share the same backbone for 3DMatch and KITTI with much fewer parameters.

\textbf{Superpoint Matching}. Following GeoTrans \cite{qin2022geometric}, we first use a linear projection to compress the feature dimension of superpoints to 192 and 96 for 3DMatch and KITTI datasets. Then, we iteratively use the geometric self-attention module and cross-attention module 3 times with 4 attention heads. Finally, another linear project is used to project the features to 192 and 128 dimensions for 3DMatch and KITTI, respectively.


\section{Detailed Introduction of RandomCrop}
\label{augmentation}

\begin{figure}
    \centering
    \includegraphics[width=0.6\textwidth]{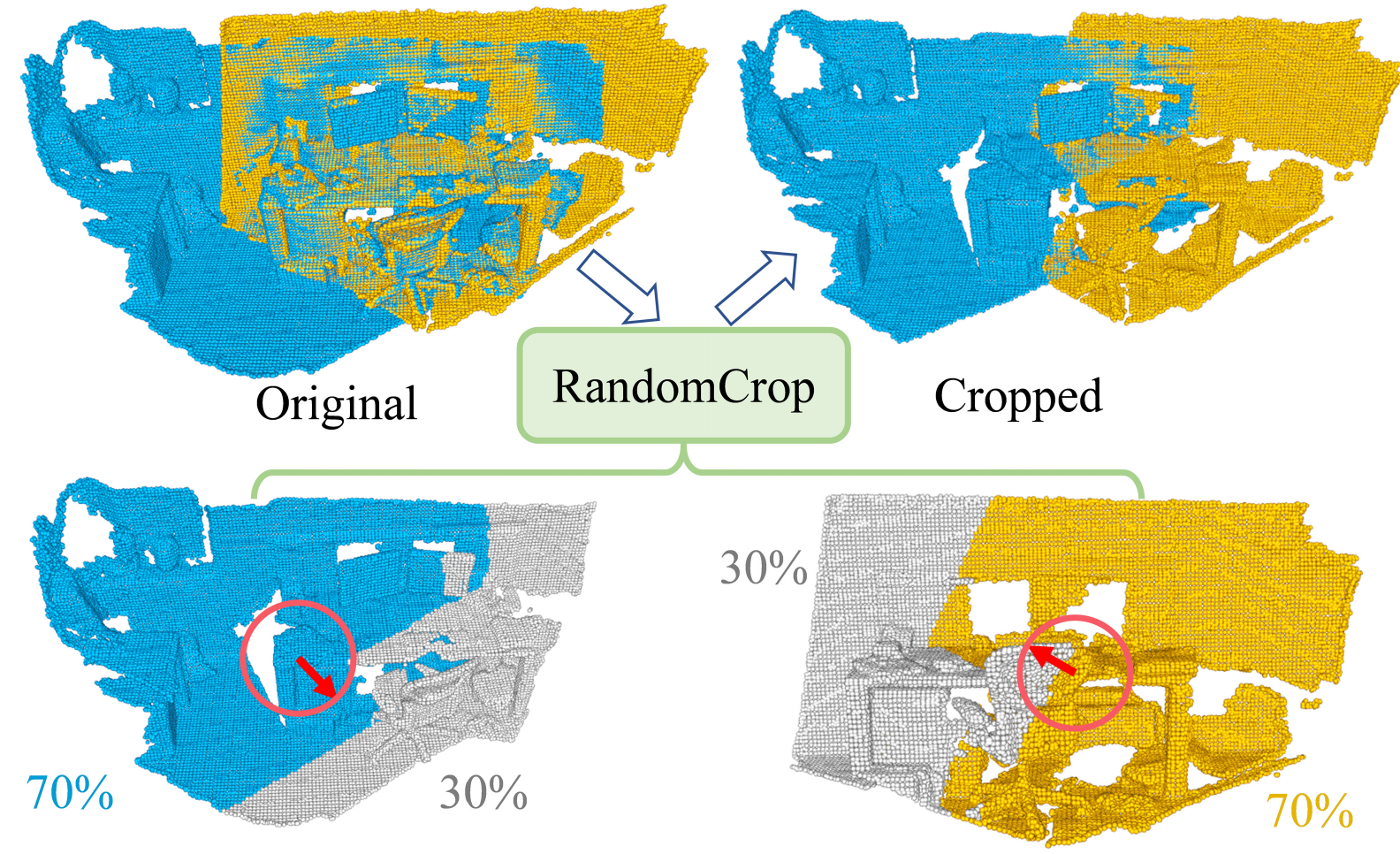}
    \caption{Diagram of RandomCrop.}
    \label{fig:aug}
\end{figure}

\begin{figure}
    \centering
    \includegraphics[width=0.7\textwidth]{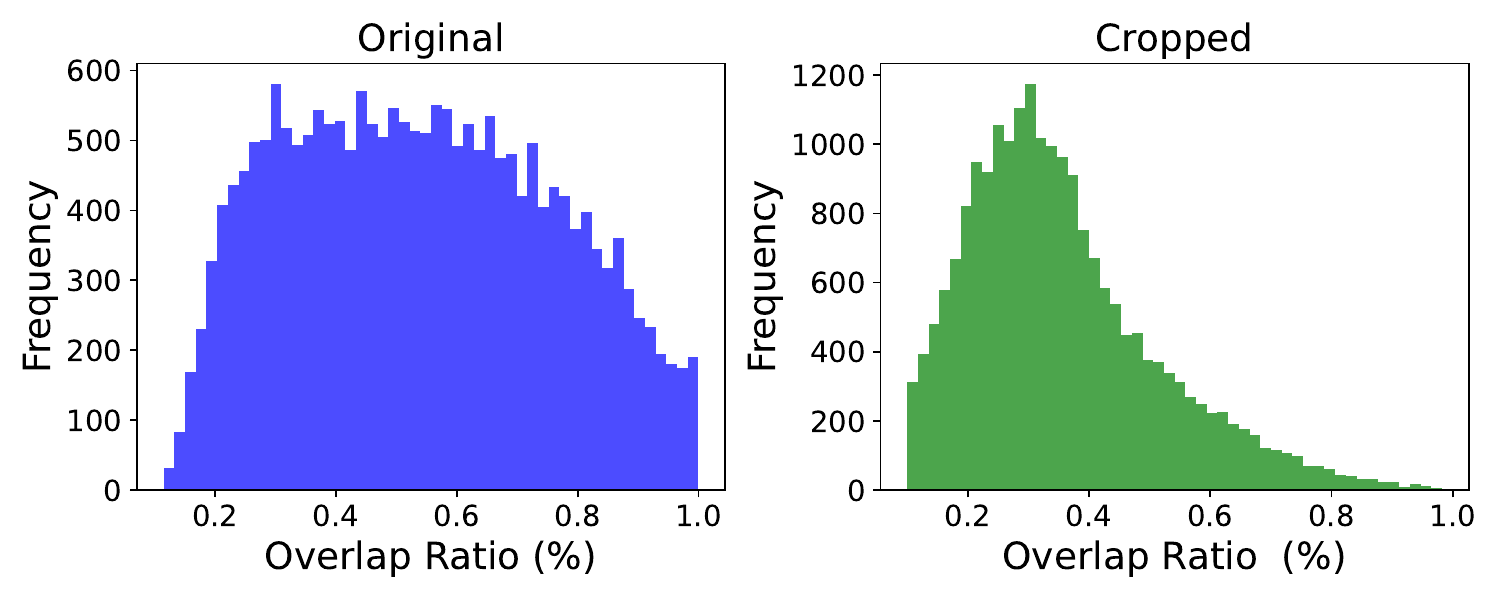}
    \caption{Distributions of overlap ratios on the original and cropped training dataset.}
    \label{fig:distribution}
\end{figure}

Geometric Transformer module \cite{qin2022geometric} utilizes cross attention to reason the global contextual information between two point clouds. Inevitably, this leads to the module being highly sensitive to the overlap distribution of the point cloud pairs. The performance of the model may degrade when testing the model on low-overlapped datasets, such as 3DLoMatch \cite{predator}. To address this issue, we present a data augmentation method, RandomCrop, to make the model more robust against low-overlapped registration.

As shown in Fig. \ref{fig:aug}, given two frames of point clouds, we first randomly generate two unit direction vectors to obtain two clipping planes perpendicular to these direction vectors. The clipping planes can divide the point clouds into two parts. We discard the part of the point clouds with a higher overlap ratio. Here, we set a hyperparameter to control the ratio of the cropped part, and empirically we set it to 0.3. As shown in Fig. \ref{fig:distribution}, we plot the overlap distributions of the 3DMatch training dataset. This illustrates that RandomCrop can reduce the overlap ratios of the training set, making the model more robust to the low-overlapped dataset.  

We conduct ablation studies to analyze the impact of RandomCrop. As shown in Table \ref{tab:ablation}, when utilizing RandomCrop, both the performance of our method and GeoTrans \cite{qin2022geometric} has been improved, especially on the 3DLoMatch dataset. These experimental results confirm the effectiveness of this data augmentation technique. 

\begin{table}[htbp]
  \centering
  \caption{Ablation experiments about RandomCrop.} 
  \label{tab:ablation}
  \begin{tabular}{l|ccl|ccl}
    \toprule
    \multirow{2}{*}{Method} & \multicolumn{3}{c|}{3DMatch} & \multicolumn{3}{c}{3DLoMatch} \\
    \cline{2-7}
     &FMR (\% $\uparrow$) & IR (\% $\uparrow$)  &RR (\% $\uparrow$) & FMR  (\% $\uparrow$)& IR (\% $\uparrow$)  &RR (\% $\uparrow$) \\
    \hline
    GeoTrans \cite{qin2022geometric} & 98.1 & 70.9 & 92.4  & 87.4 & 43.5 & 74.3 \\
    GeoTrans \cite{qin2022geometric} + R.C. &\textbf{98.7}&70.1&93.1$^{+0.7}$ &\textbf{90.8}&44.8&77.9$^{+3.6}$  \\
    Ours &98.4&76.2&93.2 &87.8&44.5&76.4\\
    Ours + R.C. & 98.5 & \textbf{76.9} &\textbf{95.0}$^{+1.8}$   & 88.3 & \textbf{47.5} & \textbf{80.5}$^{+4.1}$   \\
    \bottomrule
\end{tabular}
\end{table}

\section{Implementation}
\label{implentation}
We implement our PARE-Net with PyTorch \cite{paszke2019pytorch} on an RTX 3090 GPU with Intel (R) Xeon (R) Silver 4314 CPU. We train it with an Adam optimizer \cite{kingma2014adam}, and the detailed configurations are reported in Table \ref{tab:configuration}.

\section{Evaluation Metrics}
\label{metrics}
Following previous works \cite{predator, qin2022geometric, roreg}, we use multiple metrics to evaluate our method, including Inlier Ratio (IR), Feature Matching Recall (FMR), Inlier Ratio (IR), Registration Recall (RR), Rotation Error (RE), Translation Error (TE) and Transformation Recall (TR).

\textbf{Inlier Ratio (IR)} is the fraction of inliers among the estimated correspondences between two point clouds. A correspondence is defined as an inlier if its residual error of two points is smaller than a threshold  $\tau_{ir}$ under the ground-truth transformation $(\mathbf{R}_{gt}, \mathbf{t}_{gt})$:
\begin{equation}
    \text{IR} = \frac{1}{|\mathcal{C}|}\sum_{(\mathbf{p}_{x_i}, \mathbf{q}_{y_i}) \in \mathcal{C}}\mathds{1}\left( \|\mathbf{R}_{gt}\mathbf{p}_{x_i}+\mathbf{t}_{gt} - \mathbf{q}_{y_i}\|_2 < \tau_{ir} \right ),
\end{equation}
where $\tau_{ir}=0.1m$ and $\mathds{1}$ is the indicator function.

\textbf{Feature Matching Recall (FMR)} is the fraction of point cloud pairs whose IR is greater than a threshold $\tau_{fmr}$:
\begin{equation}
    \text{FMR} = \frac{1}{M}\sum_{i=1}^M\mathds{1}\left( \text{IR}_i > \tau_{fmr}  \right ),
\end{equation}
where $\tau_{fmr}=0.05$ and $M$ is the number of point cloud pairs to be aligned.

\textbf{Registration Recall (RR)} is the fraction of successfully aligned point cloud pairs whose root mean square error (RMSE) of ground-truth correspondences is smaller than a threshold $\tau_{rr}$ under the estimated transformation:
\begin{equation}
    \text{RR} = \frac{1}{M}\sum_{i=1}^M\mathds{1}\left( \text{RMSE}_i < \tau_{rr} \right ),
\end{equation}
where $\tau_{rr}=0.2m$ and the RMSE is computed as:
\begin{equation}
    \text{RMSE} =\sqrt{\frac{1}{|\mathcal{C}_{gt}|} \sum_{(\mathbf{p}_{x_i}, \mathbf{q}_{y_i}) \in \mathcal{C}_{gt}}\left( \|\mathbf{R}_{est}\mathbf{p}_{x_i}+\mathbf{t}_{est} - \mathbf{q}_{y_i}\|^2_2  \right )},
\end{equation}
where  $(\mathbf{R}_{est}, \mathbf{t}_{est})$ is the estimated transformation.

\textbf{Rotation Error (RE)} is the geodesic distance in degrees between ground-truth and estimated rotation matrices:
\begin{equation}
    \text{RE}= {arccos \left( \frac{\text{tr}({\bf{R}}_{est}^{-1}{\bf{R}}_{gt})-1}{2} \right)},
\end{equation}
where $\text{tr}(\cdot)$ is the trace of matrix.

\textbf{Translation Error (TE)} is the Euclidean distance between ground-truth and estimated translation vectors:
\begin{equation}
    \text{TE} = \Vert {{\bf t}_{est}-{\bf t}_{gt}} \Vert_2.
\end{equation}
Note that we compute the mean RE and mean TE of successfully aligned point cloud pairs, instead of all the point cloud pairs. This can more properly reflect the registration accuracy of different methods.

\textbf{Transformation Recall (TR)} is the fraction of successfully aligned point cloud pairs whose RE and TE are smaller than two thresholds:
\begin{equation}
    \text{TR} = \frac{1}{M}\sum_{i=1}^M\mathds{1}\left( \text{RE}_i < \tau_{r} \text{ and } \text{TE}_i < \tau_{t}\right ),
\end{equation}
where $\tau_{r}=15^\circ$, $\tau_{t}=0.3m$ and $\tau_{r}=5^\circ$, $\tau_{t}=2m$ for 3DMatch and KITTI, respectively.

\begin{table}[htbp]
    \centering
     \caption{Detailed configurations of our method.}
    \begin{tabular}{l|cc}
    \hline
    &3DMatch & KITTI  \\
    \hline
    \multicolumn{3}{c}{Training} \\
    \hline
    Batch Size & 1 & 1 \\
    Initial Learning Rate & 1e-4 & 1e-4\\
    Epoch &40 & 100\\
    Weight Decay & 1e-6 & 1e-6\\
    Learning Rate Decay &0.95 & 0.95\\
    Decay step & 1 & 4\\
    \hline
    \multicolumn{3}{c}{Data Augmentation} \\
    \hline
    Voxel Size & 0.025${m}$ & 0.3${m}$ \\
    Gaussian noise & 0.005${m}$ & 0.01${m}$ \\
    Rotation & 2$\pi$ & 2$\pi$ \\
    Scale          & - & [0.8, 1.1] \\
    Translation    & - & 2m \\
    Crop Ratio & 0.3 & 0.3 \\
      \hline
    \multicolumn{3}{c}{Network} \\
      \hline
    Number of Nearest Neighbors & 35 & 35\\
    Number of Weight Matrices & 4 & 4 \\
    Number of Coarse Correspondences $N_c$ & 256 & 256 \\
    Number of Fine Correspondences $N_f$ & 1K & 1K \\
    Acceptance Radius $\tau_d$ & 0.1${m}$ & 0.6${m}$ \\
    \hline
    \end{tabular}
    \label{tab:configuration}
\end{table}

\section{Additional Quantitative Results}
\label{quan_results}
\subsection{Comparison with Robust Transformation Estimators}
Recently, some robust transformation estimators \cite{choy2020deep, lee2021deep, bai2021pointdsc, chen2022sc2} have been proposed to generate reliable hypotheses more efficiently. They \cite{bai2021pointdsc, chen2022sc2} usually leverage spatial consistency to identify inliers of correspondences that are established by off-the-shelf descriptors, such as FCGF \cite{fcgf} and Predator \cite{predator}. Compared with RANSAC \cite{ransac}, they are more efficient and more robust to outliers. We compare our method with them on 3DMatch and 3DLoMatch. Following their protocols \cite{choy2020deep}, we use three metrics, including RE, TE, and TR. The results are reported in Table \ref{tab:robustestimator}. Note that the TR is computed by averaging all the point cloud pairs, different from the results reported in Table 2 in the main paper, which compute the scene-wise averages.

First of all, our method significantly outperforms these robust transformation estimation methods. Our method surpasses the state-of-the-art method SC$^2$-PCR \cite{chen2022sc2} by 3.4\%/13\% on 3DMatch/3DLoMatch, demonstrating the superiority of our method. Second, to demonstrate the superiority of our feature-based hypothesis proposer, we replace it with SC$^2$-PCR. When combined with SC$^2$-PCR, it performs similarly to our hypothesis proposer with 0.5\% improvement on 3DMatch but 1.2\% decrease on 3DLoMatch in terms of RR. This may be because our method only requires one correspondence to estimate the transformation, while SC$^2$-PCR needs several correspondences. Therefore, when the inlier ratio of correspondences is low, the probability of producing reliable solutions will decrease. The results demonstrate that our simple hypothesis proposer can match and even surpass the well-designed transformation estimators.

\begin{table}[htbp]
  \centering
  \caption{Comparison results with robust transformation estimators.}
  \tabcolsep=0.06cm
   \resizebox{\linewidth}{!}{
  \begin{tabular}{l|ccc|ccc}
    \hline
    \multirow{2}{*}{Method} & \multicolumn{3}{c|}{3DMatch} & \multicolumn{3}{c}{3DLoMatch} \\
      \cline{2-7}
     &  RE ($^\circ$) & TE (${cm}$)  &TR (\%)  & RE ($^\circ$) & TE (${cm}$)  &TR (\%)  \\
     \hline
     FCGF \cite{fcgf} +  DGR \cite{choy2020deep} & 2.82 & 8.36 & 78.6 & 4.17 & 10.82 & 43.8  \\
     FCGF \cite{fcgf}  +  DHVR \cite{lee2021deep}& 2.25 & 7.08 & 91.9 & 4.14 & 12.56 & 54.4\\
     FCGF \cite{fcgf}  +  PointDSC \cite{bai2021pointdsc} & 2.06 & 6.55 & 93.3 &3.87 &10.39 & 56.1\\
     FCGF \cite{fcgf}  +  SC$^2$-PCR \cite{chen2022sc2} & 2.08 & 6.55 & 93.3 & 3.77 & 10.46 &  57.8\\
     Predator \cite{predator} + DGR \cite{choy2020deep}& - & - & - &3.19 &10.01& 59.5\\
     Predator \cite{predator}  + DHVR \cite{lee2021deep}& - & - & - &4.97& 12.33& 65.4\\
     Predator \cite{predator}  + PointDSC \cite{bai2021pointdsc}& - & - & - &3.43 &9.60 & 68.9 \\
     Predator \cite{predator}  + SC$^2$-PCR \cite{chen2022sc2}& - & - & - & 3.46 & 9.58 & 69.5\\
     Ours + SC$^2$-PCR \cite{chen2022sc2}& \textbf{1.68} & \textbf{5.43} & \textbf{97.2} & \textbf{2.56} & 9.35 & 81.3\\
     Ours &1.92 & 5.55 & 96.7 &2.95 & \textbf{8.58} & \textbf{82.5}\\
    \hline
    \end{tabular}}
    \label{tab:robustestimator}
\end{table}

\subsection{Detailed Module Cost}
We report the model size and runtime of each module of our method in Table \ref{tab:component}. The coarse matching includes the geometric transformer module and superpoint matching step. The hypothesis generation contains the point matching step and the hypothesis proposal step. It can be seen that our backbone is much lighter than GeoTrans \cite{qin2022geometric} and PEAL \cite{peal}, which use KPConv \cite{kpconv} with 6.01MB parameters for 3DMatch and 24.3MB parameters for 3DLoMatch. Moreover, it also can be seen that our hypothesis proposer is computationally cheap. 

\begin{table}[htbp]
    \centering
    \caption{Detailed running times and model sizes of the components of our method. We report the mean running time overall point cloud pairs.}
    \begin{tabular}{l|cc|cc}
    \hline
    \multirow{2}{*}{Component} & \multicolumn{2}{c}{3DMatch} & \multicolumn{2}{c}{KITTI} \\
    \cline{2-5}
     &Time (s)& Size (MB)& Time (s)& Size (MB)\\
     \hline
     Data Loader& 0.05 & - &0.06 &- \\
     Backbone & 0.07& 1.51 &0.08 &1.46\\
     Coarse Matching & 0.02 & 2.26 &0.03 & 0.61\\
     Hypothesis Generation &0.03 & 0.07 &0.04 &0.01\\
     Total & 0.17 & 3.84 &0.21 &2.08\\
    \hline
    \end{tabular}
    \label{tab:component}
\end{table}

\subsection{Impact of Overlap}
\begin{figure}
    \centering
    \includegraphics[width=0.4\textwidth]{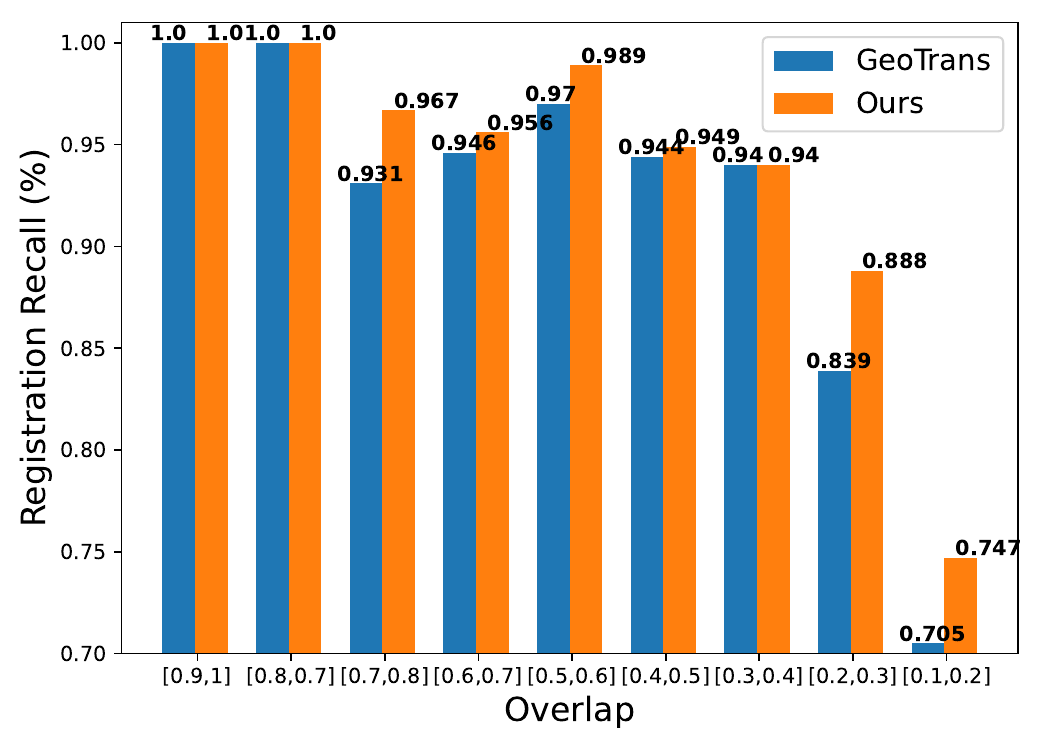}
    \caption{Comparison of our method with GeoTrans\cite{qin2022geometric} under different overlap ratios. The experimental results are reported on the union of 3DMatch and 3DLoMatch. }
    \label{fig:rr}
\end{figure}

\begin{table*}[htbp]
\caption{Comparison results of scenes on 3DMatch and 3DLoMatch. }
\setlength{\tabcolsep}{1.3mm}
\resizebox{\linewidth}{!}{
\begin{tabular}{l|rrrrrrrrr|rrrrrrrrr}
\hline
\multirow{3}{*}{Model}&\multicolumn{9}{c|}{3DMatch}&\multicolumn{9}{c}{3DLoMatch} \\
&Kitchen&Home1&Home2&Hotel1&Hotel2&Hotel3&Study&Lab&Mean&Kitchen&Home1&Home2&Hotel1&Hotel2&Hotel3&Study&Lab&Mean\\
\hline
\multicolumn{19}{c}{Registration Recall(\%$\uparrow$)} \\
\hline
PerfectMatch \cite{gojcic2019perfect}&90.6&90.6&65.4&89.6&82.1&80.8&68.4&60.0&78.4
&51.4&25.9&44.1&41.1&30.7&36.6&14.0&20.3&33.0\\

FCGF \cite{fcgf}& 98.0&94.3&68.6&96.7&91.0& 84.6&76.1&71.1&85.1
&60.8&42.2&53.6&53.1&38.0&26.8&16.1&30.4&40.1\\

D3Feat \cite{d3feat} &96.0&86.8&67.3&90.7&88.5&80.8&78.2&64.4&81.6
&49.7&37.2&47.3&47.8&36.5&31.7&15.7&31.9&37.2\\

Predator \cite{predator}&97.6&97.2&74.8&98.9&96.2&88.5&85.9&73.3&89.0
&71.5&58.2&60.8&77.5&64.2&61.0&45.8&39.1&59.8\\

CoFiNet \cite{CoFiNet}&96.4&99.1&73.6&95.6&91.0&84.6&89.7&84.4&89.3
&76.7&66.7&64.0&81.3&65.0&63.4&53.4&69.6&67.5\\

GeoTrans \cite{qin2022geometric}&98.4&97.2&83.0&97.8&92.3&88.5&90.6&91.1&92.4
&86.5&72.8&70.7&89.0&69.6&73.8&56.1&75.7&74.3\\

PEAL \cite{peal}&\bf{99.6}&\bf{98.1}&\bf{83.6}&\bf{99.5}&\bf{96.2}&\bf{96.2}&\bf{91.5}&88.9&94.2
&90.3&\bf{75.1}&\bf{76.1}&94.0&\bf{79.7}&75.5&63.7&75.7&78.8\\

Ours&\bf{99.6}&97.2&83.0&\bf{99.5}&\bf{96.2}&\bf{96.2}&91.0&\bf{97.8}&\bf{95.0}
&\bf{91.8}&73.1&73.4&\bf{94.8}&79.0&\bf{78.6}&\bf{66.2}&\bf{87.1}&\bf{80.5}\\
\hline
\multicolumn{19}{c}{Rotation Error(${^\circ}\downarrow$)}\\
 \hline
PerfectMatch \cite{gojcic2019perfect}&1.926&1.843&2.324&2.041&1.952&2.908&2.296&2.301&2.199
&3.020&3.898&3.427&3.196&3.217&3.328&4.325&3.814&3.528\\

FCGF \cite{fcgf}& \bf{1.767}&1.849&2.210&1.867&1.667&2.417&\bf{2.024}&1.792&1.949
&2.904&3.229&3.277&2.768&2.801&2.822&3.372&4.006&3.147\\

D3Feat \cite{d3feat} &2.016&2.029&2.425&1.990&1.967&2.400&2.346&2.115&2.161
&3.226&3.492&3.373&3.330&3.165&2.972&3.708&3.619&3.361\\

Predator \cite{predator}&1.861&1.806&2.473&2.045&1.600&2.458&2.067&1.926&2.029
&3.079&2.637&3.220&\bf{2.694}&2.907&3.390&\bf{3.046}&3.412&3.048\\

CoFiNet \cite{CoFiNet}&1.910&1.835&2.316&1.767&1.753&1.639&2.527&2.345&2.011
&3.213&3.119&3.711&2.842&2.897&3.194&4.126&3.138&3.280\\

GeoTrans \cite{qin2022geometric}&2.460&1.523&\bf{2.043}&\bf{1.572}&\bf{1.528}&1.691&2.149&1.978&1.868
&3.246&2.646&3.156&2.769&2.741&2.719&3.338&3.387&3.000\\

PEAL \cite{peal}&1.798&\bf{1.488}&2.071&1.580&1.592&1.704&2.077&\bf{1.672}&\bf{1.748}
&3.178&\bf{2.533}&\bf{3.031}&2.814&\bf{2.597}&\bf{2.378}&3.062&\bf{3.005}&\bf{2.825}\\

Ours   &2.363&1.510&2.242&1.621&1.577&\bf{1.542}&2.147&2.100&1.888
&\bf{3.113}&2.667&3.919&2.871&2.804&2.597&3.241&3.459&3.084\\
\hline
\multicolumn{19}{c}{Translation Error(${m}\downarrow$)}\\
 \hline
PerfectMatch \cite{gojcic2019perfect}&0.059&0.070&0.079&0.065&0.074&0.062&0.093&0.065&0.071
&0.082&0.098&0.096&0.101&\bf{0.080}&0.089&0.158&0.120&0.103\\

FCGF \cite{fcgf}& 0.053&0.056&0.071&0.062&0.061& 0.055&0.082&0.090&0.066
&0.084&0.097&0.076&0.101&0.084&0.077&0.144&0.140&0.100\\

D3Feat \cite{d3feat} &0.055&0.065&0.080&0.064&0.078&0.049&0.083&0.064&0.067
&0.088&0.101&0.086&0.099&0.092&0.075&0.146&0.135&0.103\\

Predator \cite{predator}&0.048&0.055&0.070&0.073&0.060&0.065&0.080&\bf{0.063}&0.064
&0.081&0.080&0.084&0.099&0.096&0.077&\bf{0.101}&0.130&0.093\\

CoFiNet \cite{CoFiNet}&0.047&0.059&\bf{0.063}&0.063&\bf{0.058}&\bf{0.044}&0.087&0.075&0.062
&0.080&\bf{0.078}&\bf{0.078}&0.099&0.086&0.077&0.131&0.123&0.094\\

GeoTrans \cite{qin2022geometric}&0.047&0.053&0.079&\bf{0.057}&0.061&0.051&0.081&0.079&0.064
&0.071&0.089&0.089&\bf{0.091}&0.090&0.063&0.117&0.106&0.090\\

PEAL \cite{peal}& 0.045&\bf{0.050}&0.080&0.059&0.062&0.056&\bf{0.077}&0.069&0.062
&\bf{0.069}&0.082&0.089&0.094&0.088&\bf{0.056}&0.109&\bf{0.094}&\bf{0.085}\\

Ours&\bf{0.043}&\bf{0.050}&0.080&0.058&0.060&0.048&0.078&0.075&\bf{0.061}
&\bf{0.069}&0.087&0.090&0.097&0.093&\bf{0.056}&0.109&0.122&0.090\\
\hline
\end{tabular}}
\label{tab:detailedscene}
\end{table*}

We demonstrate the experimental results of our method and GeoTrans \cite{qin2022geometric} under different overlap ratios on 3DMatch and 3DLoMatch in Fig. \ref{fig:rr}. Our method outperforms GeoTrans at different overlap ratios, especially at low overlap ratios. Our method surpasses it by 4.9\% and 4.2\% when the overlap ratios are in the [0.2, 0.3] and [0.1, 0.2] intervals, respectively. This demonstrates the robustness of our method against low-overlapped point cloud pairs.

\subsection{Scene-wise Experimental Results}

We report the scene-wise experimental results on 3DMatch and 3DLoMatch in Table \ref{tab:detailedscene}. We can see that our method achieves a relatively high RR, especially in the challenging scenario \textit{Lab}, where it outperforms the state-of-the-art method PEAL by 8.8\% and 11.4\% on 3DMatch and 3DLoMatch respectively. In terms of accuracy, our method is slightly lower than PEAL, which may be because it iteratively optimizes the transformation.

\subsection{More Ablation Study about PARE-Conv}
\label{sec:pare_ablation}
We compare the PARE-Conv with KPConv \cite{kpconv} to demonstrate the superiority of PARE-Conv. Since KPConv cannot output rotation-equivariant features, we use LGR \cite{qin2022geometric} as the transformation estimator for both PARE-Conv and KPConv. The experimental results are shown in Table \ref{tab:pare_ablation}. We can observe that PARE-Conv can generate more distinctive descriptors because its IR is significantly higher than the IR of KPConv. As a result, the RR is also improved by PARE-Conv. These experimental results strongly confirm the superiority of PARE-Conv.

\begin{table}[htbp]
  \centering
  \caption{Comparison results between PARE-Conv and KPConv \cite{kpconv}.} 
  \label{tab:pare_ablation}
  \resizebox{0.98\linewidth}{!}{
  \begin{tabular}{l|ccl|ccl}
    \toprule
    \multirow{2}{*}{Method} & \multicolumn{3}{c|}{3DMatch} & \multicolumn{3}{c}{3DLoMatch} \\
    \cline{2-7}
     &FMR (\% $\uparrow$) & IR (\% $\uparrow$)  &RR (\% $\uparrow$) & FMR  (\% $\uparrow$)& IR (\% $\uparrow$)  &RR (\% $\uparrow$) \\
    \hline
    KPConv \cite{kpconv} + LGR \cite{qin2022geometric} &98.2&69.6&92.8&87.8&43.6&76.6  \\
    PARE-Conv + LGR \cite{qin2022geometric} &\textbf{98.5}&\textbf{76.9}&\textbf{94.0}$^{+1.2}$&\textbf{88.3}&\textbf{47.5}&\textbf{78.9} $^{+2.3}$\\  
    \bottomrule
\end{tabular}}
\end{table}

\subsection{Ablation about Rotation Augmentation}
It is interesting to explore the role of rotation augmentation for rotation-equivariant networks because it theoretically has no meaning for such networks. In practice, rotation augmentation benefits PARE-Net as shown in Table \ref{tab:rotation_aug}. Because PARE-Conv can only guarantee approximate equivariant output due to computation errors. Thus, rotation augmentation can enhance the diversity of training data and is beneficial for training a better model. Moreover, we can see that rotation-sensitive GeoTrans is extremely sensitive to rotations without rotation augmentation when training, while PARE-Net is robust to rotations due to a strong bias of rotation equivariance.

\begin{table}[h]
\caption{Ablation of rotation augmentation on 3DLoMatch. We train both two methods with RandomCrop and noise.}
\centering
\begin{tabular}{cccccccc}
\hline
Train&Test&GeoTrans&PARE-Net&Train&Test&GeoTrans&PARE-Net\\
\hline
w/o & w/o & 77.5 & 77.9  & w & w/o & 75.4 & 81.3  \\
w/o & w & 11.4 & 77.8 & w & w & 72.6 & 81.8\\
\hline
\end{tabular}

\label{tab:rotation_aug}
\end{table}

\subsection{Generalization Study}
\label{sec:generalization}
\begin{table}[htbp]
  \centering
  \caption{Generalization results from 3DMatch to KITTI.}
  \begin{tabular}{l|ccc}
    \hline
     &  RE ($^\circ$) & TE (${m}$)  &TR (\%)   \\
     \hline
     Coarse-to-fine Matching & 0.79 & 0.26 & 70.8 \\
     Only Fine Matching &  0.82 & 0.19 & 98.4\\
    \hline
    \end{tabular}
    \label{tab:generalization}
\end{table}

We investigate the generalization ability of our model by directly using the model trained on 3DMatch to test on KITTI dataset. To apply the model trained on indoor small-scale scenes to outdoor large-scale point clouds, we proportionally scale the point clouds of large-scale scenes based on the voxel sizes of the two scenes. The results are shown in Table \ref{tab:generalization}. We find that the results of our approach, following the coarse-to-fine matching manner, are not satisfying, with a TR of only 70.8\%. We are puzzled by these results, as our rotation-equivariant network is lightweight and provides a strong inductive bias of rotation equivariance, which should result in good generalization capability. We speculate that the poor generalization might be attributed to the Geometric Transformer module, as it learns contextual information of the point clouds and encodes distance information in self-attention, both of which undergo significant changes in large-scale scenes. To verify this speculation, we remove the coarse matching stage and estimate the transformation directly using the point features. We randomly sample 5000 points for the two point clouds and use RANSAC to estimate the transformation, resulting in a significant improvement in TR to 98.4\%, confirming our speculation and demonstrating the good generalization performance of our PARE-Conv.

\section{Additional Qualitative Results}
\label{qual_results}
\begin{figure*}
    \centering
    \includegraphics[width=1\textwidth]{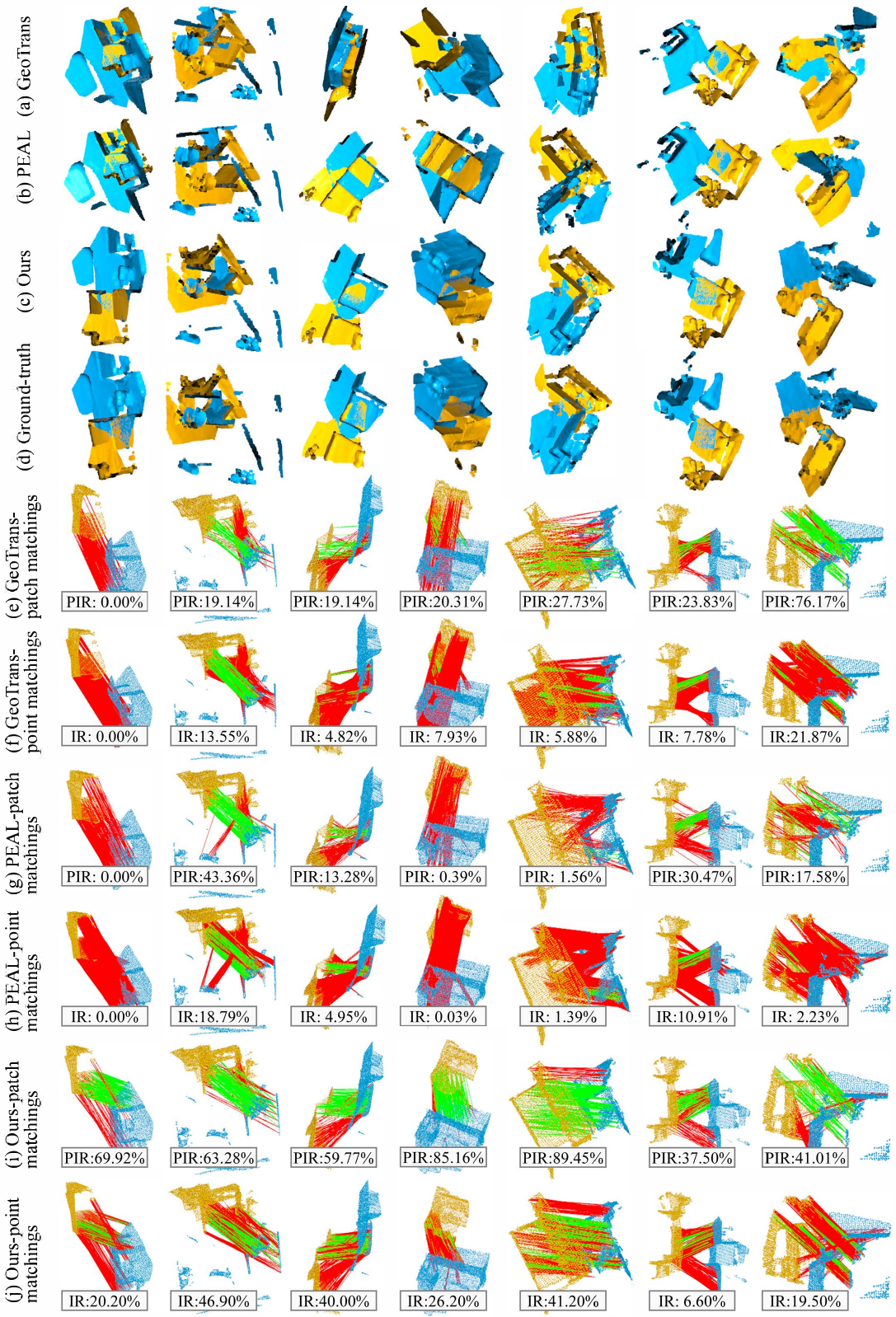}
    \caption{Visualized registration results of our method, GeoTrans \cite{qin2022geometric}, and PEAL \cite{peal}.}
    \label{fig:vis}
\end{figure*}

\begin{figure*}
    \centering
    \includegraphics[width=0.99\textwidth]{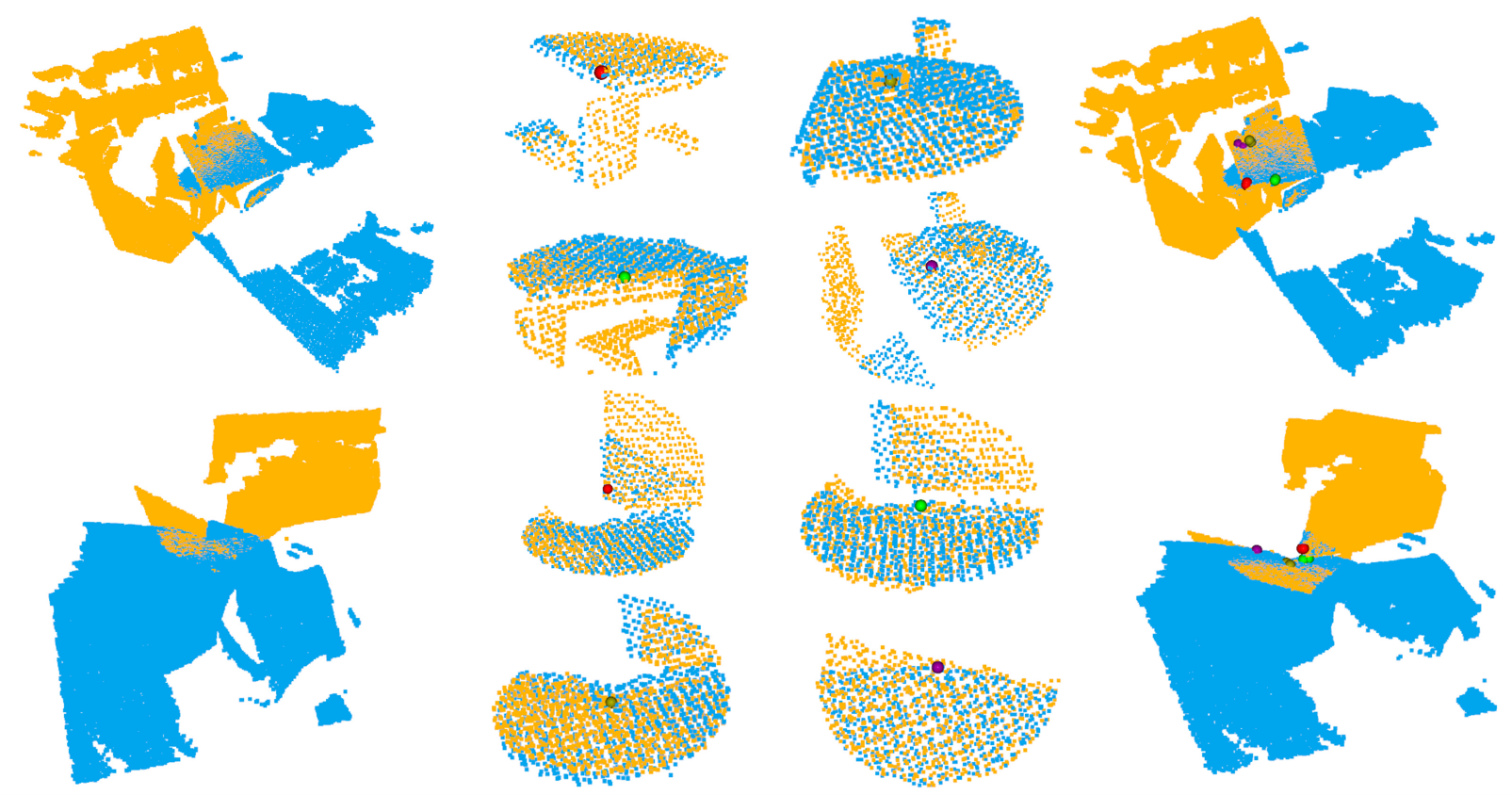}
    \includegraphics[width=0.99\textwidth]{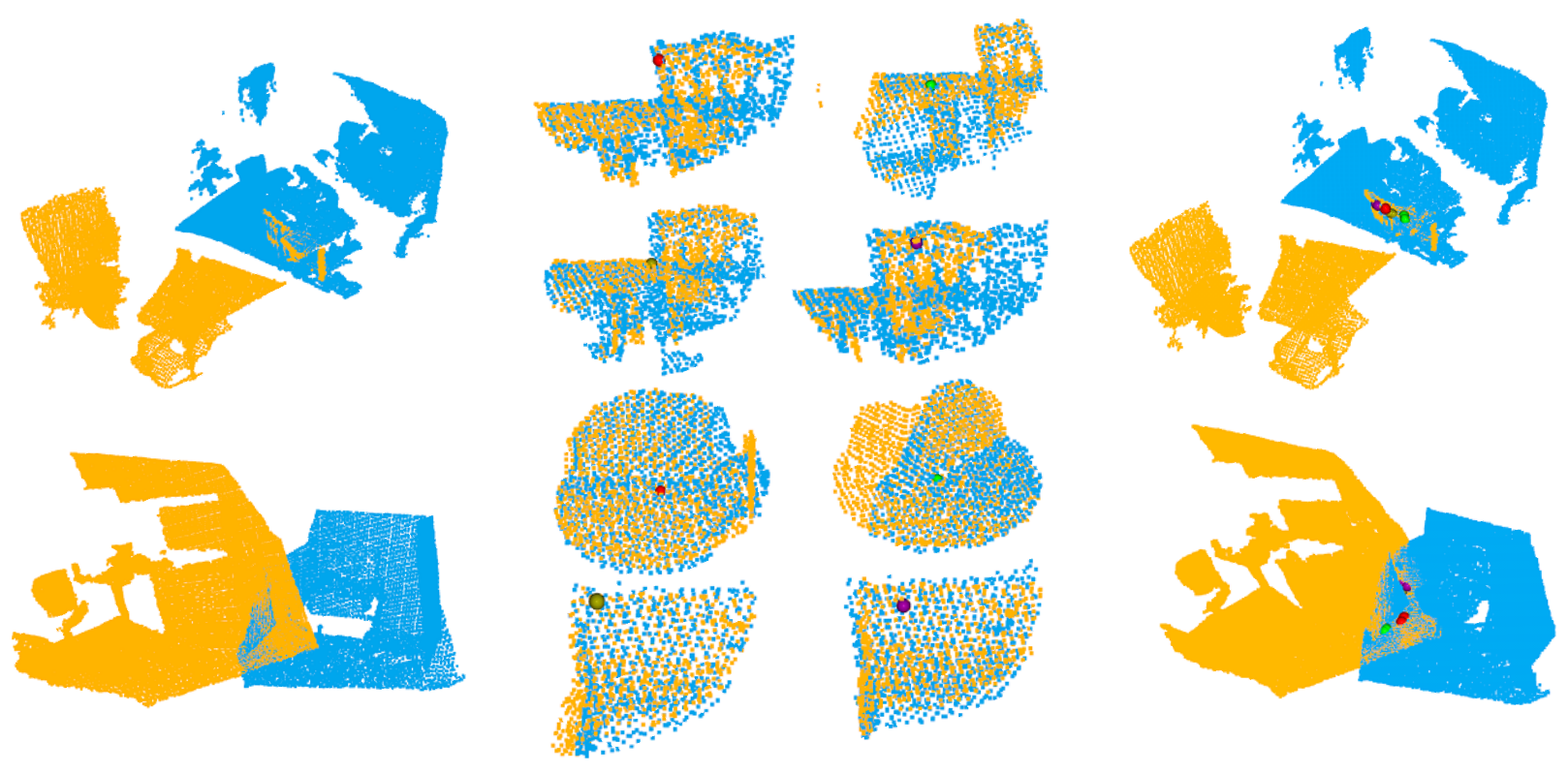}
    \includegraphics[width=0.99\textwidth]{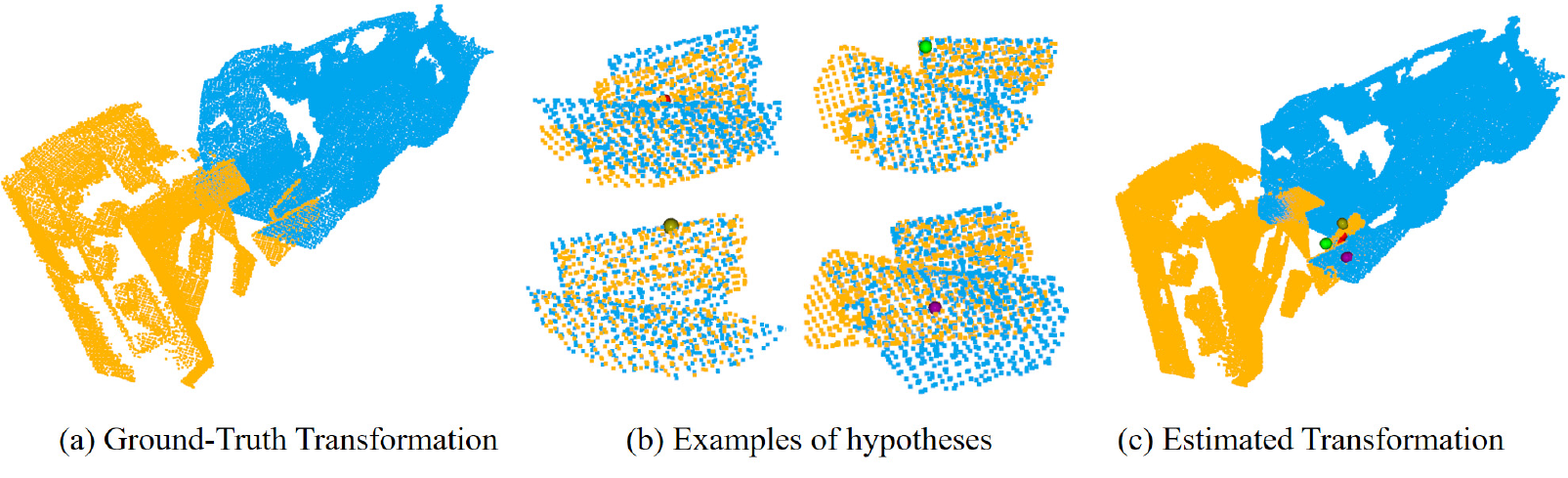}
    \caption{Visualization of proposed hypotheses on extremely low-overlapped point cloud pairs. We use four correspondences to generate four hypotheses. The corresponding points are represented by spheres of the same color. We align the local point clouds utilizing the generated hypotheses to demonstrate the accuracy of the hypotheses.}
    \label{fig:local_transforms}
\end{figure*}

We show more visualized registration results in Fig \ref{fig:vis}. Our method is able to align some point cloud pairs without obvious structural constraints, as seen in the third and sixth columns, while PEAL \cite{peal} and GeoTrans \cite{qin2022geometric} symmetrically align the point clouds. Because our rotation-equivariant features encode the directional information of the structure, which can avoid these symmetric incorrect alignments. Moreover, in Fig. \ref{fig:local_transforms}, we present some registration results of the hypotheses generated by rotation equivariant features. It can be seen that our feature-based hypothesis proposer can produce reliable solutions even when the overlap ratio is very low, demonstrating its superiority.

\section{Limitations and Future Work}
\label{limitation}
In order to better address the low-overlapped registration problem, we adopted a coarse-to-fine matching framework. However, the generalization ability is limited by the Geometric Transformer module as discussed in Sec. \ref{sec:generalization}. To reduce the ambiguity of the matching, it introduces geometric clues by encoding the relative positional relationships between superpoints, including angles and distances. However, when the scale of the point cloud changes, the distances between the superpoints also undergo significant variations, leading to a sharp degradation in the module's performance. In the future, we will further investigate more robust position encodings, such as designing relative distance encoding, to enhance its generalization performance.

%
%

\end{document}